\documentclass[a4paper,fleqn]{cas-sc}

\usepackage[numbers,sort&compress]{natbib}
\usepackage{xurl} 
\usepackage[version = 4]{mhchem}
\usepackage{placeins}
\usepackage{bm}
\usepackage{amsthm}
\newtheorem{proposition}{Proposition}


\def\tsc#1{\csdef{#1}{\textsc{\lowercase{#1}}\xspace}}
\tsc{WGM}
\tsc{QE}

\begin{document}
	\let\WriteBookmarks\relax
	\renewcommand{\topfraction}{0.92}
	\renewcommand{\bottomfraction}{0.88}
	\renewcommand{\textfraction}{0.08}
	\renewcommand{\floatpagefraction}{0.80}
	\setcounter{topnumber}{4}
	\setcounter{bottomnumber}{3}
	\setcounter{totalnumber}{6}
	\setlength{\textfloatsep}{12pt plus 2pt minus 2pt}
	\setlength{\floatsep}{10pt plus 2pt minus 2pt}
	\setlength{\intextsep}{10pt plus 2pt minus 2pt}
	
	\shorttitle{PDE-Structured Transport Operator}
	\shortauthors{Li et~al.}
	
	\title [mode = title]{PSTO: A PDE-structured transport operator with shared implicit updates for multiscale convection-diffusion-reaction systems}
	
	\author[1]{Kaiwen Li}
	\author[1]{Dezhi Zhou}
	\cormark[1]
	\ead{dezhi.zhou@sjtu.edu.cn}
    \author[2]{Kun Wu}
	\author[3]{Xingcai Lu}
	
	\affiliation[1]{organization={Global College, Shanghai Jiao Tong University},
		city={Shanghai},
		postcode={200240},
		country={China}}
	\affiliation[2]{organization={State Key Laboratory of High Temperature Gas Dynamics, Institute of Mechanics, Chinese Academy of Sciences},
    city={Beijing},
	postcode={100190},
	country={China}}
	\affiliation[3]{organization={Key Laboratory for Power Machinery and Engineering, Ministry of Education, Shanghai Jiao Tong University},
		city={Shanghai},
		postcode={200240},
		country={China}}
	
	\cortext[cor1]{Corresponding author}
	
	\begin{abstract}
		Multiscale convection--diffusion--reaction (CDR) systems couple spatial transport with local stiff reaction kinetics, making their numerical simulation computationally demanding.
		Conventional neural surrogates require costly coupled CDR training data, and their parameter counts scale with the number of transported variables, limiting scalability to high-dimensional CDR systems.
		Here we propose a PDE-Structured Transport Operator (PSTO) that learns convection--diffusion dynamics solely from reaction-free data. By embedding the PDE structure into the neural architecture, PSTO generalizes to full CDR systems at inference without retraining. A single transport operator block is shared across all transported variables and refined through relaxed fixed-point iterations, performing an implicit-like correction in latent space. This improves stability during recursive multi-step prediction without adding trainable parameters.
        We evaluate PSTO on three benchmarks: a canonical 2D CDR system, a multiscale Robertson counterflow flame, and a multiscale hydrogen planar jet flame. In every case, PSTO reduces the mean relative \(L_2\) error after 100 coupled steps by at least 50\% compared to the strongest neural baseline.
        Ablation studies on the fixed-point iterations and detailed error decomposition further confirm that PSTO achieves both high one-step accuracy and long-term stability in coupled rollouts.
        Time-step sensitivity analysis shows that PSTO maintains accuracy even with a tenfold larger time step, whereas baseline accuracies decay rapidly as the time step increases.
		This framework enables efficient, modular modeling of multiscale CDR dynamics by separating transport learning from the treatment of local kinetics.
	\end{abstract}
	
	\begin{graphicalabstract}
		\includegraphics[width=\linewidth]{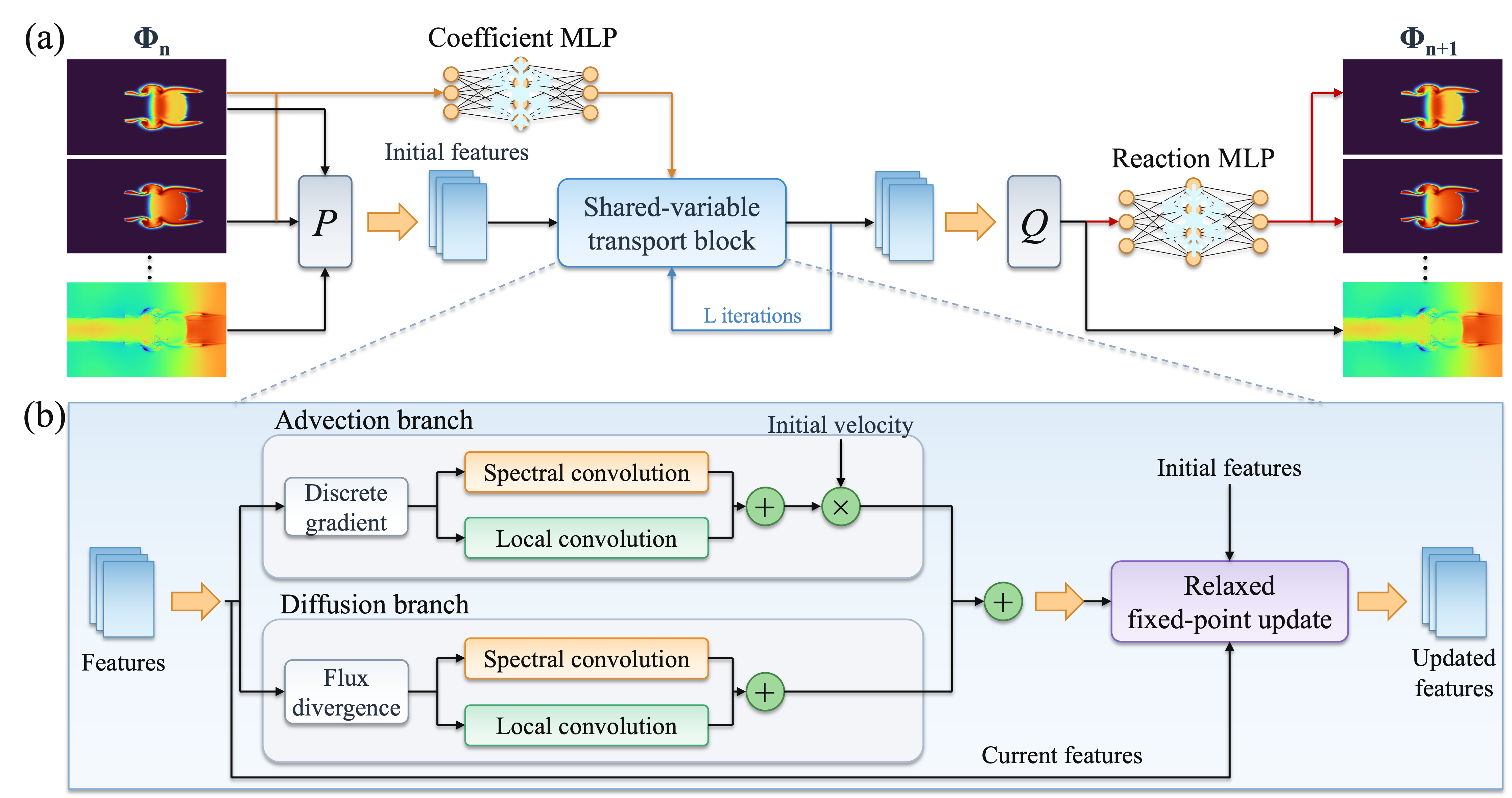}
	\end{graphicalabstract}
	\begin{highlights}
		\item By learning transport dynamics from reaction-free data alone, PSTO removes the requirement for expensive coupled CDR training trajectories, enabling seamless integration with existing numerical reaction solvers at inference.
		
		\item A PDE-structured convection--diffusion block is conditioned on velocity, density, and variable-specific transport coefficients and shared across physical variables, enabling parameter-efficient scaling to high-dimensional CDR systems.
		
		\item Relaxed fixed-point iterations reuse the transport block to approximate a learned implicit update in latent space, improving prediction accuracy and stability through iterative refinement without adding trainable parameters.
		
		\item Tested on one canonical and two multiscale CDR benchmarks, PSTO consistently reduces the mean variable-wise relative $L_2$ error after 100 coupled steps by more than 50\% compared with the best-performing neural baseline in each case.
	\end{highlights}
	
	\maketitle

	\section{Introduction}
	\label{sec:intro}
	
	Convection--diffusion--reaction (CDR) equations describe processes ranging from turbulent combustion and atmospheric dispersion to biomedical transport. In this context, transport denotes the spatial evolution of state variables through convection and diffusion, whereas reaction describes their local production and consumption. A reduced, non-conservative form of the governing equation for a CDR system is
	\begin{equation}
		\label{eq:cdr_template}
		\frac{\partial \bm{\phi}}{\partial t} + \mathbf{u}\cdot\nabla\bm{\phi} = \rho^{-1}\nabla\cdot \left( \bm{a}\nabla\bm{\phi} \right) + \dot{\bm{r}}(\bm{\phi}),
	\end{equation}
	where $\bm{\phi}$ is the state vector, $\mathbf{u}$ is the velocity field, $\rho$ is the density, $\bm{a}$ contains the variable-specific coefficients, and $\dot{\bm{r}}$ contains the local reaction source terms. These time-dependent partial differential equations (PDEs)~\cite{larsson2003partial,burger2020implicit} are conventionally solved using numerical methods. However, the computational complexity caused by the chemical stiffness and disparate time scales of CDR dynamics~\cite{young1977numerical,najm1998semi}, often makes accurate, real-time simulation of high-dimensional, multiscale CDR systems difficult or even computationally prohibitive.
	
	Recent advances in data storage and computing hardware have paved the way for deep learning as an efficient alternative for modeling complex CDR systems~\cite{lecun2015deep}. Deep learning has shown promising applications in solving high-dimensional PDEs~\cite{han2018solving}, estimating nonlinear systems~\cite{haber2018stable}, and discovering dynamical representations~\cite{lusch2018deep}. Among neural surrogate architectures, neural operator frameworks have become increasingly popular because of their strong capabilities in PDE modeling. Representative architectures include the Deep Operator Network (DeepONet)~\cite{lu2021learning} and the Fourier Neural Operator (FNO)~\cite{li2020fourier}. With its ability to capture nonlocal spatial interactions through learned Fourier kernels, FNO has been applied to various engineering problems, including plasma modeling~\cite{gopakumar2023fourier} and stiff chemical kinetics~\cite{weng2025extended}. FNO has also been extended to improve its performance in specific modeling tasks. For example, F-FNO improves purely data-driven modeling through novel representations~\cite{tran2021factorized}. IS--FNO uses reversibility and the structure of spectral evolution to improve long-horizon predictions~\cite{yu2026inverse}. U-FNO enriches spatial representations with local convolutional branches~\cite{wen2022u}, while MscaleFNO rescales its inputs to learn high-frequency mappings~\cite{you2025mscalefno}. Despite these advances, neural operators applied to multiscale CDR systems still face a major limitation: the lack of explicit physical knowledge makes long autoregressive rollouts prone to error accumulation under stiff kinetics and sharp gradients. 
	
	To address these limitations, physics-informed and physics-embedded learning approaches have been developed~\cite{karniadakis2021physics}. Physics-informed neural networks~\cite{raissi2019physics}, physics-informed mixture-of-experts models~\cite{sharma2025physics}, and physics-constrained surrogates for flow and reaction kinetics~\cite{zhu2019physics,zhang2024crk} use governing equations to guide model training. Physical constraints can also be imposed during inference to enforce conservation laws and nonlinear equalities~\cite{utkarsh2026physics}. A complementary strategy embeds physical principles directly into neural architectures, as exemplified by PDE-Net~\cite{long2018pde,long2019pde}, deep operator splitting networks~\cite{lan2023dosnet}, physics-encoded recurrent convolutional neural networks~\cite{rao2023encoding}, physics-enhanced deep surrogates~\cite{pestourie2023physics}, and PDE-preserved neural networks~\cite{liu2024multi}. For modeling spatial transport, NeuroDDAF incorporates wind-dependent convection--diffusion dynamics in Fourier space and combines the resulting prediction with a learned forecast~\cite{dey2026neuroddaf}. In coupled PDE systems, architectural design must additionally account for interactions among the evolving physical fields. COMPOL represents these interactions by aggregating process-specific latent features and passing the resulting information to the constituent neural operators~\cite{qu2025compol}. Studies on coupled FNO variants introduce interactions between fields in Fourier space and examine how shared and field-specific operator components affect predictions of coupled dynamics~\cite{jing2026extending}. These studies provide architectural approaches for learning interactions among physical fields, with applications to canonical cases such as Darcy flow and Gray--Scott reaction--diffusion systems. Nevertheless, accurate modeling and effective optimization remain challenging for high-dimensional CDR systems with coupled, multiscale dynamics. The presence of disparate time scales and stiffness can destabilize autoregressive predictions when neural surrogates employ explicit time advancement within a residual learning framework~\cite{he2016deep}. These difficulties have motivated two complementary strategies for improving stability and efficiency in multiscale CDR modeling: operator splitting and implicit time integration.
	
	Operator splitting is a widely used strategy for decoupling multiscale dynamics by decomposing a coupled problem into subproblems, solving them separately, and combining the resulting updates~\cite{strang1968construction,ren2014dynamic}. The resulting splitting error depends on the time-step size and the order of the splitting scheme. In CDR systems, this decomposition enables transport and reaction computations to be accelerated independently. Previous efforts have primarily focused on accelerating the computationally dominant reaction process: several studies replace local reaction solvers with neural surrogates~\cite{owoyele2022chemnode,kumar2024combustion}, while others accelerate stiff ordinary differential equation integration on graphics processing units (GPUs)~\cite{zhang2024graphics}. Coupling these accelerated reaction solvers with conventional transport solvers can substantially reduce simulation costs~\cite{wang2025enforcing}. As reaction costs decrease, transport accounts for a larger share of the total runtime. For instance, in the Cambridge stratified-flame case, the fraction of runtime spent on chemistry decreases from 85.8\% to 37.6\% after acceleration~\cite{zhang2026performance}. In the EPICC air quality model, accelerating gas-phase chemistry reduces its runtime share from 45.7\% to 29.9\%~\cite{cao2026enhancing}. GPU-accelerated transport solvers have therefore been developed~\cite{wang2025gpu}, but their performance remains sensitive to memory access patterns, load imbalance, and communication overhead~\cite{carreon2025gpu}. Alternatively, neural surrogates can accelerate transport through fast, efficient inference and can be coupled with accelerated reaction solvers to provide complementary acceleration of the full CDR simulation. This approach has considerable potential for highly efficient surrogate modeling, but strategies for training transport surrogates at low computational cost while retaining high physical interpretability remain largely unexplored.
	
	Implicit time integration offers a strategy for improving numerical stability and relaxing the time-step restrictions of explicit schemes in stiff, multiscale CDR simulations. These advantages have motivated the development of implicit neural surrogates. For instance, the implicit physics-integrated differentiable model (Im-PiNDiff) advances spatiotemporal states through a neural implicit equation~\cite{akhare2025implicit}, but requires iterative solvers and careful convergence control. IFNO interprets shared Fourier layers as fixed-point iterations~\cite{you2022learning}, but may still involve substantial parameter counts and training data requirements. Consequently, these frameworks may incur a substantial computational burden and require large models and extensive training data to obtain converged solutions at each time step.
	
	Overall, existing neural surrogates still face challenges in simultaneously achieving physical consistency, data efficiency, modeling simplicity, and autoregressive stability. To address these challenges, we propose a PDE-Structured Transport Operator (PSTO). First, PSTO learns transport independently and couples with a separate reaction solver to infer the full multiscale CDR system. This separation decouples the dynamics and substantially reduces training data costs. Second, the PSTO architecture is constructed from the governing transport equations and explicitly conditioned on velocity and variable-specific transport coefficients, directly enhancing its physical interpretability and robustness. Third, PSTO is shared across different physical variables, limiting parameter growth as the dimensionality of the CDR system increases. Fourth, the PSTO transport block is applied repeatedly through a finite number of relaxed fixed-point iterations to approximate a learned implicit update in latent space. This design aims to achieve implicit-like stability in prediction while avoiding the computationally demanding iterations required for fully converged implicit evolution.
	
	The remaining paper is structured as follows. Section~\ref{sec:psto_method} introduces the governing system and describes the overall architecture, the theoretical basis, the local models, and the baseline transport surrogates. Section~\ref{sec:error_analysis} derives error bounds for the coupled system, and Section~\ref{sec:numerical_methodology} details the training and evaluation procedures. Section~\ref{sec:results} presents the numerical experiments, while Section~\ref{sec:conclusion} summarizes the main findings and outlines future work.

	\section{PDE-Structured Transport Operator}
	\label{sec:psto_method}
	
	\subsection{Governing equations of multiscale CDR systems}
	\label{sec:governing_equations}
	
	Multiscale CDR systems couple momentum transport, heat transfer, and species evolution across disparate spatial and temporal scales~\cite{tomboulides1997numerical,nonaka2018conservative}. The physical state typically comprises the velocity $\mathbf{u}$, pressure $p$, temperature $T$, and species mass fractions $Y_i$. Density $\rho$ and transport properties enter the governing equations as coefficients and may depend on the local thermochemical state. We consider the reduced governing system
	\begin{equation}
		\left\{
		\begin{aligned}
			\partial_t\rho+\nabla\cdot(\rho\mathbf{u})&=0,\\
			\partial_t\mathbf{u}+\mathbf{u}\cdot\nabla\mathbf{u} &=\rho^{-1}\nabla\cdot(a_u\nabla\mathbf{u})-\rho^{-1}\nabla p + \dot{r}_{\mathbf{u}},\\
			\partial_tT+\mathbf{u}\cdot\nabla T &=\rho^{-1}\nabla\cdot(a_T\nabla T)+\dot{r}_T,\\
			\partial_tY_i+\mathbf{u}\cdot\nabla Y_i &=\rho^{-1}\nabla\cdot(a_{Y_i}\nabla Y_i)+\dot{r}_{Y_i}, \qquad i=1,2,\ldots N_s.
		\end{aligned}
		\right.
		\label{eq:multiscale_governing_equations}
	\end{equation}
	The transport coefficients are $a_u=\mu$, $a_T=\lambda/c_p$, and $a_{Y_i}=\rho D_i=a_T/Le_i$, where $\mu$ is the dynamic viscosity, $\lambda$ is the thermal conductivity, $c_p$ is the specific heat capacity, $D_i$ is the species diffusivity, $N_s$ is the number of species, and $Le_i$ is the species Lewis number. Here, $p$ denotes pressure, and the source terms $\dot{r}$ represent local production and consumption, with $\dot{r}_{\mathbf{u}}=0$ unless a velocity source is prescribed. Componentwise viscous diffusion is adopted for weakly dilatational flows with slowly varying viscosity. The temperature equation neglects pressure work and viscous heating for small pressure changes relative to the thermal energy scale and low Brinkman numbers, respectively. Weak $c_p$ gradients permit the stated conductive form, while the species-diffusion contribution to temperature evolution is neglected. Species transport follows Fick’s law, omitting thermal, pressure, and cross-diffusion. In the low-Mach-number formulation, hydrodynamic pressure represents a small perturbation to the thermodynamic pressure and satisfies an elliptic equation enforcing consistency between velocity and density evolution~\cite{day2000numerical}. We therefore omit an additional pressure output and include pressure effects in the velocity update.
	
	With thermophysical properties treated as state-dependent coefficients and pressure effects incorporated through the learned velocity update, we define the predicted state as $\Phi=(\mathbf{u},T,Y_1,\ldots,Y_{N_s})$ and express Eq.~\eqref{eq:multiscale_governing_equations} in an operator form as
	\begin{equation}
		\partial_t\Phi=\mathcal T(\Phi)+\mathcal R(\Phi),
	\end{equation}
	where $\mathcal T$ and $\mathcal R$ denote the spatial transport and local reaction operators, respectively. Let $\mathcal T_{\Delta t}$ and $\mathcal R_{\Delta t}$ denote the finite-time solution maps generated by these operators over a time interval $\Delta t$. The Lie--Trotter and Strang operator splitting schemes are then given by
	\begin{equation}
		\mathcal S^{\rm LT}_{\Delta t} =\mathcal R_{\Delta t}\circ\mathcal T_{\Delta t}, \qquad \mathcal S^{\rm St}_{\Delta t} =\mathcal R_{\Delta t/2}\circ\mathcal T_{\Delta t}\circ\mathcal R_{\Delta t/2}.
		\label{eq:splitting}
	\end{equation}
	This decomposition allows transport and reaction to be accelerated separately and their surrogate models to be trained on independently generated data. In this work, PSTO learns transport dynamics from reaction-free CD trajectories, without requiring coupled CDR data to train the transport surrogate, as described in Section~\ref{sec:PSTO}. Local constitutive relations and reaction kinetics are modeled independently using low-dimensional, spatially independent samples or evaluated using local numerical solvers, as discussed in Section~\ref{sec:local_models}.
	
	\subsection{Network architecture}
	\label{sec:PSTO}
	
	The transport terms for velocity components and scalar variables in Eq.~\eqref{eq:multiscale_governing_equations} share a convection--diffusion structure, which can be expressed using a common transport operator:
	\begin{equation}
		\mathcal T(q;h) =A(q,\mathbf u)+D(q,a,\rho).
		\label{eq:block_response}
	\end{equation}
	Here, $q$ denotes a transported variable, either a velocity component, $T$, or $Y_i$. Its transport coefficient $a$ is selected from $a_u$, $a_T$, and $a_{Y_i}$, respectively. The tuple $h=(\mathbf u,a,\rho)$ collects the conditioning fields.
	
\begin{figure}[pos=htbp]
		\centering
		\includegraphics[width=0.98\linewidth]{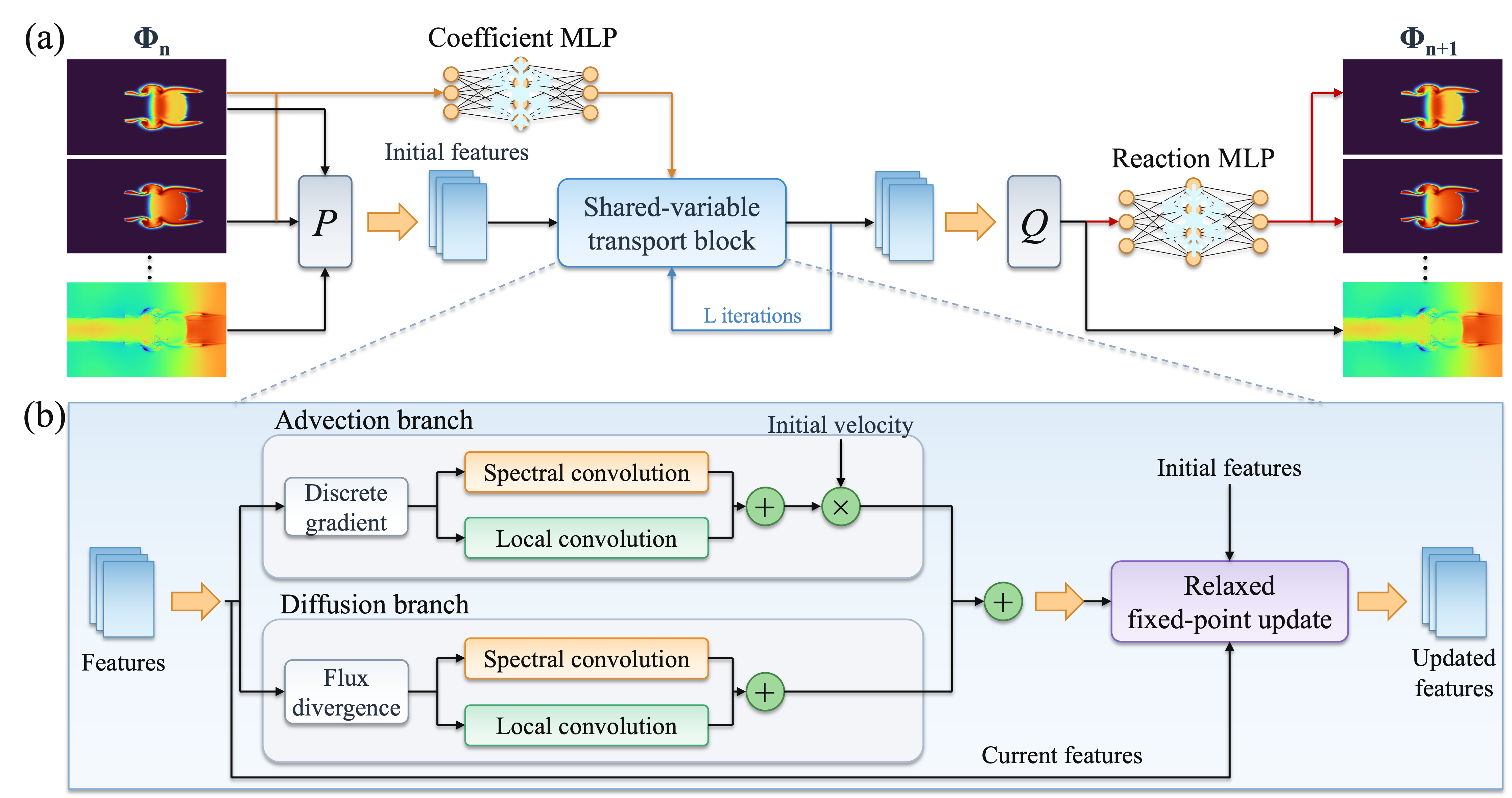}
		\caption{PSTO architecture. (a) A coupled Lie--Trotter transport--reaction prediction step with shared lifting map $P$, transport block, and projection map $Q$. The coefficient and reaction MLPs are trained independently, and their parameters remain fixed during PSTO training. Orange arrows indicate coefficient evaluation, black arrows indicate transport evolution, and red arrows indicate reaction coupling. The transport output follows Eq.~\eqref{eq:psto_grouping}, with the subtraction of $z^0$ before projection and the addition of $q^n$ afterward omitted from the schematic. (b) The conditioned advection and diffusion branches and the relaxed fixed-point update defined in Eq.~\eqref{eq:iteration_update}. The flux-divergence operation incorporates the variable-specific transport coefficient and density weighting in Eq.~\eqref{eq:block_response}. The initial features $z^0$ and the conditioning fields remain fixed throughout the $L$ latent iterations.}
		\label{fig:structure}
	\end{figure}

	Figure~\ref{fig:structure}(a) depicts the overall architecture for a single coupled update of $\Phi_n$. The framework consists of two models, PSTO (comprising a local coefficient model and a transport block shared across variables) and a local reaction model. These two models are trained separately and coupled through operator splitting according to Eq.~\eqref{eq:splitting} at inference. Based on the common structure in Eq.~\eqref{eq:block_response}, PSTO is designed as a unified transport framework, with all network components shared across variables and variable-specific behavior determined solely by the input states, coupling between variables and boundary conditions. Specifically, PSTO evolves the states in physical units and enforces the boundary conditions. It also includes a coefficient multilayer perceptron (MLP) to explicitly evaluate density and transport coefficients. This coefficient MLP can thus accommodate various cases when the thermophysical properties vary in space and time. These quantities then serve as conditioning fields during state evolution. Before the shared transport block is applied, a shared bias-free pointwise linear map $P$ first lifts each transported variable $q^n$ to a high-dimensional latent representation. The latent state is then updated and its residual is passed to a subsequent shared bias-free pointwise linear map $Q$, which projects the latent residual to a physical increment. Overall, PSTO performs the update
	\begin{equation}
		q_L=q^n+Q(z^L-Pq^n), \qquad q^n \in \Phi_n.
		\label{eq:psto_grouping}
	\end{equation}
	Here, $L$ is the number of iterations of the shared transport block. Further details of this iterative procedure are shown in Fig.~\ref{fig:structure}(b). The reaction MLP, as a representative local reaction model, advances the transported variables produced by PSTO, updating the thermochemical state while leaving velocity components without source terms unchanged. The resulting variables are concatenated to form the next state $\Phi_{n+1}$. Repeating this coupled update yields autoregressive predictions from the prescribed initial state $\Phi_0$.
	
	Figure~\ref{fig:structure}(b) then illustrates the transport block in latent space and its relaxed fixed-point update in detail. The transport block comprises advection and diffusion branches that combine spectral convolutions for nonlocal spatial interactions with local convolutions for spatially local feature transformations. In the advection branch approximating $A(q,\mathbf u)$ in Eq.~\eqref{eq:block_response}, discrete directional gradients of the latent features are multiplied by the corresponding velocity components. In the diffusion branch approximating $D(q,a,\rho)$, density and transport coefficients are incorporated into the weighted flux-divergence operation. The transport block is applied over $L$ relaxed fixed-point iterations to obtain implicit-like stability while avoiding the computational cost of a fully converged implicit update. Given the initial high-dimensional latent representation $z^0=Pq^n$ and the transport block $\mathcal T_\theta$, we define the implicit residual $F_\theta$ for the latent state $z$ after the transport evolution. At layer $\ell$, the relaxed fixed-point update produces the next latent representation $z^{\ell+1}$. Their respective expressions are given by
	\begin{align}
		F_\theta(z;q^n,h) &=z-Pq^n-\mathcal T_\theta(z;h), \\
		z^{\ell+1} &=z^\ell-\omega F_\theta(z^\ell;q^n,h) =(1-\omega)z^\ell +\omega\{z^0+\mathcal T_\theta(z^\ell;h)\}, \qquad 0 \leq \ell \leq L-1.
		\label{eq:iteration_update}
	\end{align}
	The relaxation factor is parameterized as $\omega=\tfrac12\operatorname{sigmoid}(\eta)$, where $\eta$ is a trainable scalar. Section~\ref{sec:proof_of_fixed-point} establishes an error bound for PSTO's approximation of the implicit solution.
	
	The architectural designs offer four potential advantages over conventional neural surrogates for CDR simulations.
	\begin{itemize}
		\item \textbf{Training costs can be substantially reduced.} As described in Section~\ref{sec:governing_equations}, PSTO and the local reaction models are trained using two efficiently generated datasets. The CD trajectories for PSTO are obtained from convection--diffusion simulations without reaction-kinetics coupling. Boundary conditions are specified using either the equilibrium state of a zero-dimensional reactor or a representative thermochemical state sampled from the reaction zone, facilitating efficient learning of transport dynamics across all species.
		\item \textbf{Physical consistency and generalizability are enhanced.} We construct the network architecture based on the common convection--diffusion structure, conditioning transport evolution on the evaluated physical coefficients.
		\item \textbf{The number of trainable transport parameters is independent of the number of transported variables.} Because the architecture is shared across variables, including the lifting, transport, and projection blocks, the framework scales efficiently to high-dimensional CDR systems.
		\item \textbf{Prediction stability and inference efficiency are improved.} A finite number of relaxed fixed-point iterations avoids the cost of a fully converged implicit update while improving stability relative to conventional explicit neural surrogates. The resulting stability gains can permit larger time steps, offsetting the additional cost per step and improving overall inference efficiency.
	\end{itemize}
	
	\subsection{Convergence analysis of the relaxed fixed-point update}
	\label{sec:proof_of_fixed-point}
	
	PSTO employs relaxed fixed-point iterations to approximate the implicit transport update. Proposition~\ref{prop:iteration_fixed_point} shows that this approximation converges to the exact implicit solution under mild conditions on the transport operator, and provides an explicit error bound that decays with the number of iterations.
    
	\begin{proposition}[Transport prediction error]
		\label{prop:iteration_fixed_point}
		Let $q^{n+1}=\mathcal T_{\Delta t}(q^n)$ denote the reference transport state. Under the assumptions stated in Appendix~\ref{app:supplementary_residual_estimates}, the prediction error satisfies
		\begin{equation}
			\|q_L-q^{n+1}\|_2 \leq \epsilon_{PQ} +\frac{\|Q\|_2}{1-\gamma} \left(r_L+\omega\epsilon_{\rm imp}\right).
			\label{eq:projected_implicit_accuracy}
		\end{equation}
		Here, $\gamma\in[0,1)$ is the contraction factor of the relaxed iteration, and $\|Q\|_2$ is the induced Euclidean operator norm. The residual and reconstruction terms are defined by
		\begin{equation}
			r_L =\omega\|F_\theta(z^L;q^n,h)\|_2,\quad \epsilon_{\rm imp} =\|P(q^{n+1}-q^n)-\mathcal T_\theta(Pq^{n+1};h)\|_2,\quad \epsilon_{PQ} =\|(I-QP)(q^{n+1}-q^n)\|_2.
			\label{eq:reference_residual_and_projection_error}
		\end{equation}
		The quantity $r_L$ measures the scaled residual at the final latent iterate, $\epsilon_{\rm imp}$ measures the residual of the learned implicit relation at the lifted reference state, and $\epsilon_{PQ}$ measures the reconstruction error of the reference increment under $QP$. The symbol $I$ denotes the identity map.
	\end{proposition}
	
		The proof is given in Appendix~\ref{app:supplementary_residual_estimates}.
	%
	Equation~\eqref{eq:projected_implicit_accuracy} identifies three contributions to PSTO's transport error bound. Under the contraction assumption, increasing $L$ reduces the latent iteration residual $r_L$. Training the transport operator can reduce $\epsilon_{\rm imp}$, while optimizing the lifting and projection maps can reduce the reconstruction error $\epsilon_{PQ}$.
	
	\subsection{Local coefficient and reaction models}
	\label{sec:local_models}
	
	In addition to approximating the common differential operators through a shared transport block, PSTO incorporates density and variable-specific transport coefficients as conditioning fields. When these quantities are not prescribed as constants, they depend on the local thermochemical state. We therefore use an independent local constitutive model to map the input thermochemical state to the required conditioning fields, thereby representing the physical coupling among variables during transport evolution. A local reaction solver is also applied at each grid point and coupled with the transport surrogate to advance the full CDR system.
	
	The evaluation of constitutive properties and the integration of reaction kinetics can both be performed using local numerical solvers. To accelerate the full CDR simulation, these computations can also be replaced by neural surrogates. Figure~\ref{fig:structure} illustrates the local coefficient and reaction models. Since both constitutive relations and reaction kinetics depend on local thermochemical states, we use separate MLPs to represent the corresponding pointwise mappings. The coefficient MLP predicts the instantaneous density and transport coefficients from the thermochemical state at each grid point and is evaluated at the beginning of each transport substep. The reaction MLP predicts the finite-time evolution of the local thermochemical state. Velocity components are excluded from both models because the constitutive relations and reaction kinetics considered here have no explicit velocity dependence.
	
	As described in Section~\ref{sec:governing_equations}, the local models are trained using low-dimensional samples without spatial coupling. The input thermochemical states for these samples can be obtained from one-dimensional CDR simulations or a few snapshots of the target CDR system. For each input state, zero-dimensional numerical solvers evaluate the thermophysical properties and advance the reaction kinetics over a prescribed interval. The resulting properties provide targets for the coefficient model, while the evolved thermochemical states provide targets for the reaction model. These local evaluations and integrations enable efficient generation of the input--target pairs without spatially coupled computations.
	
	Once the two local models have been pretrained, their parameters remain fixed during PSTO training and coupled inference. For a fair comparison, all transport surrogates, including the baseline models in Section~\ref{sec:baselines}, use the same reaction map and splitting composition within each experiment.
	
	\subsection{Baseline models}
	\label{sec:baselines}
	
	We evaluate PSTO against three representative neural surrogates, FNO~\cite{li2020fourier}, U-Net~\cite{ronneberger2015u}, and U-FNO~\cite{wen2022u}, spanning spectral, multiscale convolutional, and hybrid architectures. FNO captures nonlocal transport dynamics through global Fourier convolutions. It serves as a baseline to assess whether PSTO's local convolutional structure better resolves small-scale transport features. U-Net employs a multiscale convolutional encoder–decoder suited to capturing dynamics across spatial scales. Thus, U-Net tests whether PSTO's physics-conditioned design and spectral-style shared operator improve large-scale flow representation. U-FNO combines spectral and local operations, providing a closer architectural reference. Its inclusion isolates the benefits of PSTO's PDE-structured formulation and fixed-point updates in terms of physical consistency, generalization, and stability.
	
	All baselines process the physical variables jointly and predict residual increments from the same reaction-free CD training data as PSTO. Inputs are normalized to $[-1,1]$, and predictions are converted back to physical units before being composed with the common reaction map. Optimization hyperparameters, training epochs, and all other experimental settings are kept identical across PSTO and the baselines to ensure a fair comparison, isolating the effect of network architecture.
	
    \subsection{Coupled Error and autoregressive stability analysis}
	\label{sec:error_analysis}
	
    Because coupled inference relies on operator splitting between transport and reaction surrogates, we begin by decomposing the coupled prediction error into its constituents and deriving a rigorous upper bound. Building on this error analysis, we then establish stability conditions for PSTO during autoregressive rollouts. The combination of these two analyses provides a set of sufficient conditions under which PSTO guarantees both accuracy and stability in long-term predictions.
	
	\begin{proposition}[autoregressive prediction error]
		\label{prop:coupled_error_bound}
		Let $\widehat x_{n+1}=\widehat{\mathcal S}^{\chi}_{\Delta t}(\widehat x_n)$ and $x_{n+1}=\mathcal I_{\Delta t}(x_n)$ denote the approximate and exact coupled trajectories, with $\widehat x_0=x_0$ and $\|x_0\|_2>0$. Here, $\chi\in\{\rm LT,\rm St\}$ specifies Lie--Trotter or Strang splitting. Under the assumptions stated in Appendix~\ref{app:proof_coupled_error}, the normalized prediction error $e_n=\|\widehat x_n-x_n\|_2/\|x_0\|_2$ satisfies
        \begin{equation}
            e_n\leq \frac{\delta_{\Delta t}^{\chi}}{\|x_0\|_2} \sum_{j=0}^{n-1} \bigl(\Lambda_{\widehat{\mathcal S}^{\chi}}\bigr)^j, \qquad n\geq1.
            \label{eq:split_rollout_error}
        \end{equation}
		Here, $\Lambda_{\widehat{\mathcal S}^{\chi}}$ is a uniform Lipschitz constant of the approximate coupled map on $\mathcal D$, and
		\[
		\delta_{\Delta t}^{\chi}=\sup_{x\in\mathcal D}\|\widehat{\mathcal S}^{\chi}_{\Delta t}(x)-\mathcal I_{\Delta t}(x)\|_2
		\]
		is its uniform one-step error.
	\end{proposition}

    The proof is given in Appendix~\ref{app:proof_coupled_error}. Equation~\eqref{eq:split_rollout_error} shows that, when $\Lambda_{\widehat{\mathcal S}^{\chi}}<1$, the error satisfies $e_n\leq\delta_{\Delta t}^{\chi}/\|x_0\|_2 \cdot (1-\Lambda_{\widehat{\mathcal S}^{\chi}})^{-1}$ and remains uniformly bounded over the admissible rollout. When $\Lambda_{\widehat{\mathcal S}^{\chi}}=1$, the estimate reduces to $e_n\leq n\delta_{\Delta t}^{\chi}/\|x_0\|_2$, allowing at most linear error growth. For $\Lambda_{\widehat{\mathcal S}^{\chi}}>1$, the bound permits geometric growth. Thus, reducing the one-step error tightens the rollout error bound, while keeping the coupled Lipschitz constant at or below one prevents geometric amplification in this estimate.

	\section{Training and evaluation setups}
	\label{sec:numerical_methodology}
	
	\subsection{Data preprocessing}
	\label{sec:training}
	
	State variables in multiscale CDR systems differ substantially in units and magnitude. PSTO evolves fields in physical units through its shared transport block, explicitly conditioned on density and transport coefficients. The baselines process all variables jointly and use $\mathcal N_{-1,1}$ to balance their input scales. For consistent loss scaling, PSTO and all baselines apply $\mathcal N_{0,1}$ to both predicted and reference fields, reducing imbalances caused by differences in variable magnitude. The two maps are defined as
    \begin{equation}
    \mathcal N_{-1,1}(q)=2\frac{q-q_{\min}}{q_{\max}-q_{\min}}-1,\qquad \mathcal N_{0,1}(q)=\frac{q-q_{\min}}{q_{\max}-q_{\min}},
    \label{eq:field_normalization}
    \end{equation}
    where $q_{\min}$ and $q_{\max}$ denote the variable-specific normalization bounds determined from the corresponding datasets. All transport surrogates are trained on reaction-free CD trajectories and evaluated through coupled CDR rollouts.
    
    Species mass fractions can span several orders of magnitude, yet minor species may remain kinetically important. For the constitutive and reaction models, we therefore apply the Box--Cox transformation~\cite{box1964analysis},
    \begin{equation}
    f_{\rm BCT}(Y)=\frac{Y^\beta-1}{\beta},\qquad \beta>0,
    \label{eq:species_box_cox_transform}
    \end{equation}
    where an appropriate choice of $\beta$ places greater relative emphasis on variations at low mass fractions than affine scaling alone. The transformed values are then standardized using species-wise means and standard deviations from the training data, placing the species features on comparable scales.
	
	\subsection{Training objective for the coefficient and reaction models}

    For each spatially local model, let $V\in\mathbb R^{M\times d_{\rm out}}$ contain the standardized reference outputs for $M$ samples and $d_{\rm out}$ output variables, and let $\widehat V$ denote the corresponding predictions. The mean absolute error (MAE) and mean squared error (MSE) are defined as
    \begin{equation}
    \mathcal L_{\mathrm{MAE}}=\frac{1}{Md_{\mathrm{out}}}\sum_{j,k}|\widehat V_{jk}-V_{jk}|,\qquad \mathcal L_{\mathrm{MSE}}=\frac{1}{Md_{\mathrm{out}}}\sum_{j,k}(\widehat V_{jk}-V_{jk})^2.
    \label{eq:local_model_losses}
    \end{equation}
    Both the coefficient and reaction models are trained by minimizing $\mathcal L_{\mathrm{MAE}}$, whose linear penalty limits the dominance of large residuals during optimization. Both models are evaluated using $\mathcal L_{\mathrm{MSE}}$, whose quadratic penalty emphasizes large prediction errors, providing a complementary assessment of accuracy across thermochemical states.
	
    \subsection{Training objective for the transport surrogates}
	\label{sec:composite_transport_loss}
	
	The transport objective combines data losses, species constraints, and PDE residuals to balance prediction accuracy and physical consistency. The data losses fit the reference fields while emphasizing regions with strong gradients and multiscale structures. Species constraints penalize negative mass fractions and deviations from a unit sum, while PDE residuals discourage violations of the applicable transport and continuity equations. With $\bar q=\mathcal N_{0,1}(q)$ and the corresponding normalized prediction $\widehat{\bar q}$, the relative $L_2$ error averaged over variables is
	\begin{equation}
		\mathcal L_{\mathrm{rel}}=\frac{1}{C}\sum_{c=1}^{C}\frac{\|\widehat{\bar q}_c-\bar q_c\|_2}{\|\bar q_c\|_2 + \epsilon}.
		\label{eq:relative_loss}
	\end{equation}
	Here, $C$ is the number of output variables, and the norms are evaluated over the batch and spatial grid. Normalization and variable-wise averaging help balance the contributions of different fields. $\epsilon$ is a small positive constant.
	
	A region-of-interest (ROI) loss emphasizes reactive regions with strong spatial gradients,
	\begin{equation}
		\mathcal L_{\mathrm{ROI}}=\frac{1}{C}\sum_{c=1}^{C}\frac{\langle w(\widehat{\bar q}_c-\bar q_c)^2\rangle}{\langle w\bar q_c^2\rangle + \epsilon}.
		\label{eq:roi_loss}
	\end{equation}
	The weights are defined by $w=(g/\|g\|_\infty)^o$, with $g=|\nabla_h\bar q_r|$, where $q_r$ is a representative transported scalar used to identify regions with pronounced gradients. The exponent $o$ controls the degree of gradient emphasis, and the weights are normalized independently for each sample. The notation $\langle\cdot\rangle$ denotes averaging over batch samples and spatial grid points.
	
	For multiscale reconstruction, we adapt the wavelet loss of Zhu et~al.~\cite{zhu2021wavelet} using a two-level Haar decomposition. Let $W_\ell$ collect the horizontal, vertical, and diagonal detail coefficients at level $\ell$, and let $A_2$ denote the second-level low-frequency approximation. These coefficients are computed from the normalized reference fields, with hats indicating their predicted counterparts. The wavelet loss is
    \begin{equation}
    \mathcal L_{\mathrm{wave}}={}40\,\mathcal L_{\mathrm{MSE}} \left(\|\widehat W_1\|_{\mathrm{dir}},\|W_1\|_{\mathrm{dir}}\right) +10\,\mathcal L_{\mathrm{MAE}}(\widehat W_1,W_1) +2.5\,\mathcal L_{\mathrm{MAE}}(\widehat W_2,W_2) +0.2\,\mathcal L_{\mathrm{MSE}}(\widehat A_2,A_2).
    \label{eq:wavelet_loss}
    \end{equation}
    Here, $\|\cdot\|_{\mathrm{dir}}$ denotes the pointwise Euclidean norm across the three detail directions. Each MAE or MSE term is averaged over batch samples, variables, and the corresponding coefficient entries.
	
	Then the combined data objective is
	\begin{equation}
		\mathcal L_d=\mathcal L_{\mathrm{rel}}+\lambda_{\mathrm{ROI}}\mathcal L_{\mathrm{ROI}}+\lambda_{\mathrm{wave}}\mathcal L_{\mathrm{wave}}.
		\label{eq:wave_data_loss}
	\end{equation}
	
	When species mass fractions are predicted, we penalize deviations from a unit sum and negative values using
	\begin{equation}
		\mathcal L_{\mathrm{cons}}=\lambda_{\mathrm{mass}}\left\langle\left(\sum_{i=1}^{N_s}\widehat Y_i-1\right)^2\right\rangle+\frac{\lambda_{\mathrm{neg}}}{N_s}\sum_{i=1}^{N_s}\left\langle[\max(0,-\widehat Y_i)]^2\right\rangle.
		\label{eq:cons_loss}
	\end{equation}
	Here, $\widehat Y_i$ is the predicted physical mass fraction.
	
	We also include a PDE constraint to guide optimization toward physically consistent solutions. For each transported variable, the auxiliary PDE residual is defined as
	\begin{equation}
		R_{q_c}=\frac{\widehat q_c^{\,n+1}-q_c^n}{\Delta t}+\mathbf u_n\cdot\nabla\widehat q_c^{\,n+1}-\frac{1}{\rho^n}\nabla\cdot\left(a_c^n\nabla\widehat q_c^{\,n+1}\right),
		\label{eq:pde_scalar_residual}
	\end{equation}
	where the transport coefficients $a_c^n$ and density $\rho^n$ are evaluated from the input state or prescribed as constants. Both the transport and continuity residuals use the input velocity $\mathbf u_n$ as the advecting velocity. Spatial derivatives are evaluated on interior cells using centered differences for advection and constant-coefficient diffusion, and weighted flux divergences for diffusion with spatially varying coefficients. For systems involving pressure gradients, the auxiliary momentum residual is omitted because the pressure-gradient term is not included in the simplified formulation.
	
	When mass continuity is part of the governing system, we also define
	\begin{equation}
		R_{\mathrm{cont}}=\frac{\widehat\rho^{\,n+1}-\rho^n}{\Delta t}+\nabla\cdot(\widehat\rho^{\,n+1}\mathbf u_n),
		\label{eq:pde_continuity_residual}
	\end{equation}
	where $\rho^n$ and $\widehat\rho^{\,n+1}$ are reconstructed from the input and predicted thermochemical states, respectively. This density reconstruction couples the continuity penalty to the predicted thermochemical fields.
	
	To balance variables with different physical scales, each transport residual is divided by the corresponding variable range, while the continuity residual is divided by the input density. Denoting the normalized residuals by $\widetilde R_j$, the aggregate PDE loss is
	\begin{equation}
		\mathcal L_{\mathrm{PDE}}^{\mathrm{raw}}=\frac{1}{|\mathcal J|}\sum_{j\in\mathcal J}\left\langle|\widetilde R_j|^2\right\rangle,
		\label{eq:pde_raw_loss}
	\end{equation}
	where $\mathcal J$ indexes the applicable transport residuals and includes the continuity residual when mass continuity is part of the governing system. The normalized residuals have units of $\mathrm{s}^{-1}$.
	
	Overall, the complete transport objective is
	\begin{equation}
        \mathcal L = \mathcal L_d+\mathcal L_{\mathrm{cons}} +\lambda_{\mathrm{PDE}}\lambda_{\mathrm{adaptive}}^{(k)} \mathcal L_{\mathrm{PDE}}^{\mathrm{raw}},\qquad \lambda_{\mathrm{adaptive}}^{(k)}=\frac{\mathcal L_d(\theta_k)}{\mathcal L_{\mathrm{PDE}}^{\mathrm{raw}}(\theta_k)+\epsilon}.
        \label{eq:psto_total_loss}
	\end{equation}
	
	The adaptive factor is recomputed at each optimization step and detached from the computational graph. It rescales the PDE loss relative to the data loss, while the gradients continue to propagate.

	\subsection{Evaluation protocol}
	\label{sec:evaluation_protocol}
	
	Within each test case, all models start each rollout from the same randomly selected reacting CDR state. Relative errors are computed from fields in physical units as
	\begin{equation}
		\varepsilon_c(n)=\frac{\|\widehat q_c^n-q_c^n\|_2}{\|q_c^n\|_2},\qquad \varepsilon(n)=\frac1C\sum_c\varepsilon_c(n).
		\label{eq:evaluation_metric}
	\end{equation}
	The norms are evaluated over the spatial grid. Each variable's relative error is computed separately before averaging across variables. ``Ground Truth'' denotes the numerical reference solution. The evaluation steps and reported prediction horizons are specified in each case.
	
	To compare prediction horizons with flow durations, we define the reference time
	\begin{equation}
		t_{\mathrm{ref}}=\frac{L_x}{\Delta u_{\max}}.
		\label{eq:velocity_range_time}
	\end{equation}
	Here, $L_x$ is the streamwise domain length, and $\Delta u_{\max}$ denotes the streamwise velocity variation over the dataset for the corresponding case.

	\section{Numerical experiments}
	\label{sec:results}
	
	We evaluate PSTO on three CDR systems that test different aspects of the proposed framework. The canonical system in Section~\ref{sec:2d_CDR} examines transport prediction with nonlinear polynomial sources. Robertson counterflow in Section~\ref{sec:2d_robertson} introduces stiff reaction kinetics with steady velocity fields. The hydrogen planar jet flame in Section~\ref{sec:2d_planarjet} combines twelve predicted fields with varying transport properties. We train the transport surrogates on reaction-free CD trajectories and generate the CDR references by numerically solving the full governing equations. Within each case, all models use the same initial state, reaction treatment, boundary conditions, and applicable species constraints. Prediction errors are reported after 1, 5, 10, 20, 50, and 100 steps.
	
	The three cases also assess the shared transport parameterization as the number of predicted variables increases. PSTO shares its lifting, transport, and projection maps across variables, so its transport parameter count remains unchanged. Each architecture retains the same internal settings across the cases, apart from the input and output dimensions required by each problem. Table~\ref{tab:model_sizes} lists the configurations and parameter counts, and Appendix~\ref{app:training_settings} provides the training settings. Finally, Section~\ref{sec:error_propagation} evaluates the effectiveness of iterative refinement, analyzes one-step errors and their propagation during autoregressive rollouts, and examines the trade-off between predictive accuracy and inference efficiency.
	
	\subsection{Two-dimensional canonical CDR system}
	\label{sec:2d_CDR}
	
	The first case examines whether transport learned from reaction-free data remains accurate when nonlinear local sources are included in the prediction. The canonical CDR system comprises two velocity components, $\mathbf u=(u,v)$, and a scalar $\phi$ with bistable reaction kinetics. The governing equations are
	\begin{equation}
		\begin{aligned}
			u_t&=\nu\Delta u-\mathbf u\cdot\nabla u+(1-u^2-v^2)u+(u^2+v^2)v,\\
			v_t&=\nu\Delta v-\mathbf u\cdot\nabla v+(1-u^2-v^2)v-(u^2+v^2)u,\\
			\phi_t&=\nu\Delta\phi-\mathbf u\cdot\nabla\phi+\phi(1-\phi)(\phi-\tfrac12).
		\end{aligned}
		\label{eq:canonical_cdr_system}
	\end{equation}
	The velocity transport terms follow the Burgers benchmark considered in physics-encoded learning~\cite{rao2023encoding}, while the nonlinear source terms couple $u$ and $v$ locally and introduce bistable kinetics for $\phi$. The periodic unit-square domain is discretized on a $100\times100$ grid with $\Delta x=0.01\,\mathrm{m}$ and $\Delta t=2.5\times10^{-4}\,\mathrm{s}$. All three variables share the diffusivity $\nu=0.005\,\mathrm{m^2/s}$. A Python-based numerical solver generates the CDR reference data by solving the full system in Eq.~\eqref{eq:canonical_cdr_system}. The reaction-free CD training data are generated by solving the same system with the polynomial source terms omitted.
	
	All surrogate models are evaluated against the same CDR reference trajectory from the same reacting initial state. Field comparisons and spatial error maps are presented at step 100, corresponding to an elapsed time of $0.025\,\mathrm{s}$ or approximately $9.0\%$ of the reference time $t_{\mathrm{ref}}=0.279\,\mathrm{s}$.
	
	\begin{table}[pos=htbp]
		\centering
		\caption{Mean variable-wise relative $L_2$ error (\%) for the canonical CDR system, evaluated using Eq.~\eqref{eq:evaluation_metric}. Columns indicate the number of coupled steps.}
		\label{tab:rel2_2d_CDR}
		\small
		\setlength{\tabcolsep}{4pt}
		\begin{tabular}{lrrrrrr}
			\toprule
			Model & 1 & 5 & 10 & 20 & 50 & 100 \\
			\midrule
			PSTO & 0.012 & 0.056 & 0.110 & 0.211 & 0.477 & 0.893 \\
			FNO & 0.013 & 0.061 & 0.116 & 0.239 & 0.910 & 2.82 \\
			U-Net & 0.008 & 0.034 & 0.060 & 0.128 & 0.599 & 2.03 \\
			U-FNO & 0.009 & 0.039 & 0.069 & 0.145 & 0.719 & 2.49 \\
			\bottomrule
		\end{tabular}
	\end{table}
	
	Table~\ref{tab:rel2_2d_CDR} shows that U-Net achieves the lowest errors at the early evaluation steps, whereas PSTO has the lowest errors at steps 50 and 100. At step 100, PSTO reaches a mean relative error of $0.893\%$, compared with $2.03\%$ for the best-performing baseline U-Net, corresponding to a reduction of approximately $56.1\%$. The change in ranking shows that a lower one-step prediction error does not necessarily lead to a lower error after repeated coupled updates.
	
	\begin{figure}[pos=htbp]
		\centering
		\includegraphics[width=0.98\linewidth]{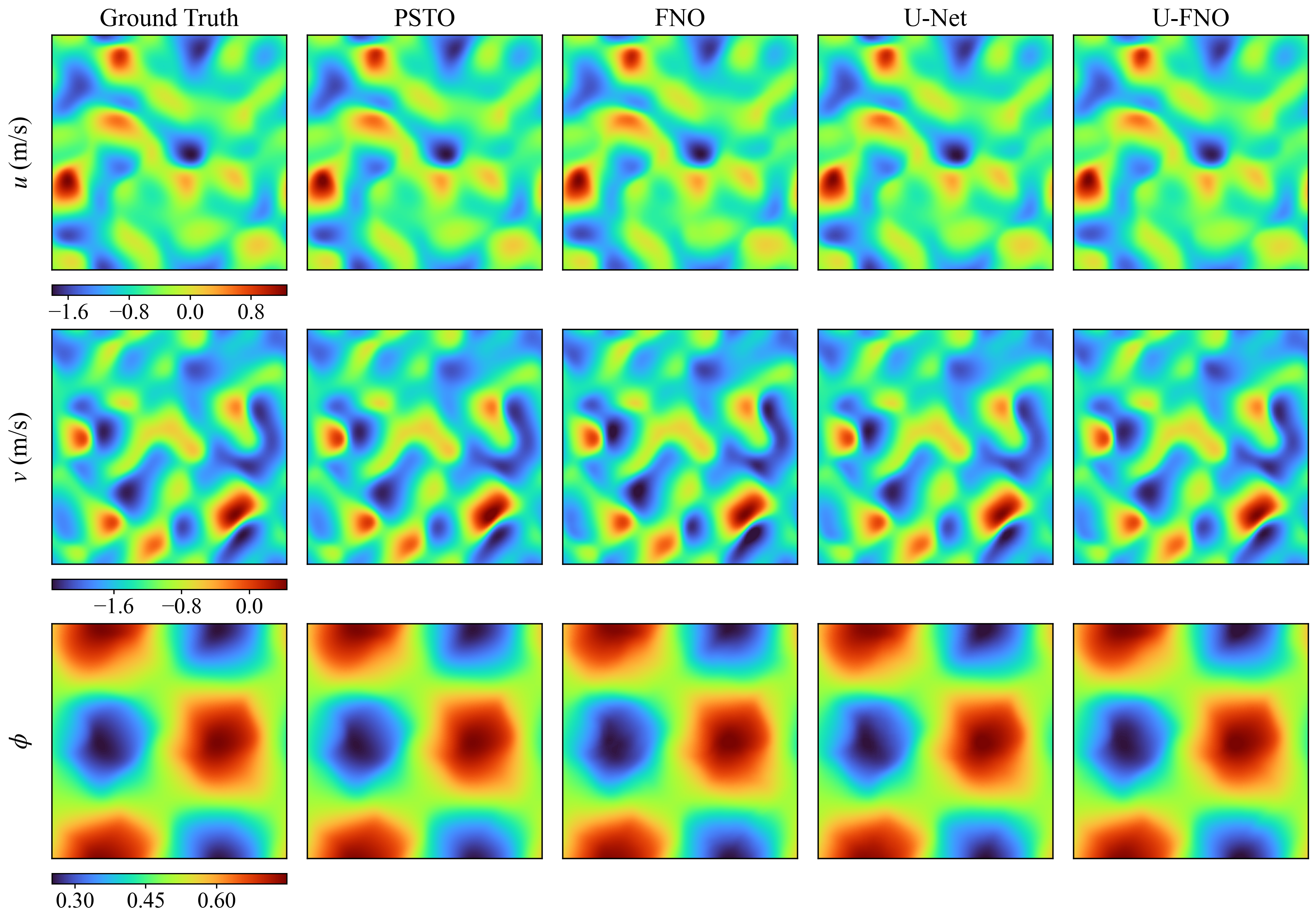}
		\caption{Canonical CDR fields after 100 coupled steps ($t-t_0=0.025\,\mathrm{s}$). Rows show $u$, $v$, and $\phi$, and columns show the numerical reference, PSTO, FNO, U-Net, and U-FNO. All surrogate models start from the same initial state and include the same prescribed polynomial source terms.}
		\label{fig:2d_CDR_results}
	\end{figure}
	
	Figure~\ref{fig:2d_CDR_results} shows that all models retain the main spatial patterns, but PSTO achieves the lowest endpoint error in every predicted field. Its $\phi$ error is $75.4\%$ lower than the best baseline error. Variable-wise errors are listed in Table~\ref{tab:channels_CDR}, and spatial error maps are provided in Appendix~\ref{app:spatial_errors}.
	
	To assess consistency with the governing dynamics in addition to agreement with the reference fields, we evaluate a discrete residual of the full CDR equations along each coupled prediction. Unlike the reaction-free residual used for training the transport surrogates in Section~\ref{sec:composite_transport_loss}, this diagnostic includes the source terms in Eq.~\eqref{eq:canonical_cdr_system}. The advecting velocity is the predicted velocity $\widehat{\mathbf u}^{\,n}$ at the beginning of the coupled step. With $\Delta q_{c,\max}=q_{c,\max}-q_{c,\min}$, the normalized residual is
	\begin{equation}
		\mathfrak r_c^n=\frac{1}{\Delta q_{c,\max}}\left[\frac{\widehat q_c^{n+1}-\widehat q_c^n}{\Delta t}+\mathcal A_h(\widehat{\mathbf u}^{\,n},\widehat q_c^{n+1})-\nu\Delta_h\widehat q_c^{n+1}-\dot{r}_c(\widehat q^{n+1})\right],\qquad \mathfrak r^n(\mathbf x)=\left(\frac{1}{3}\sum_{c=1}^3|\mathfrak r_c^n(\mathbf x)|^2\right)^{1/2}.
		\label{eq:canonical_rollout_residual}
	\end{equation}
	Here $\mathcal A_h$ and $\Delta_h$ denote the centered finite-difference advection operator and the five-point diffusion operator, respectively. The sources $\dot{r}_c$ are evaluated at the predicted next state. All trajectories begin from the same physical initial state, and the normalized residual has units of $\mathrm{s}^{-1}$. 
	
	\begin{figure}[pos=htbp]
		\centering
		\includegraphics[width=0.64\linewidth]{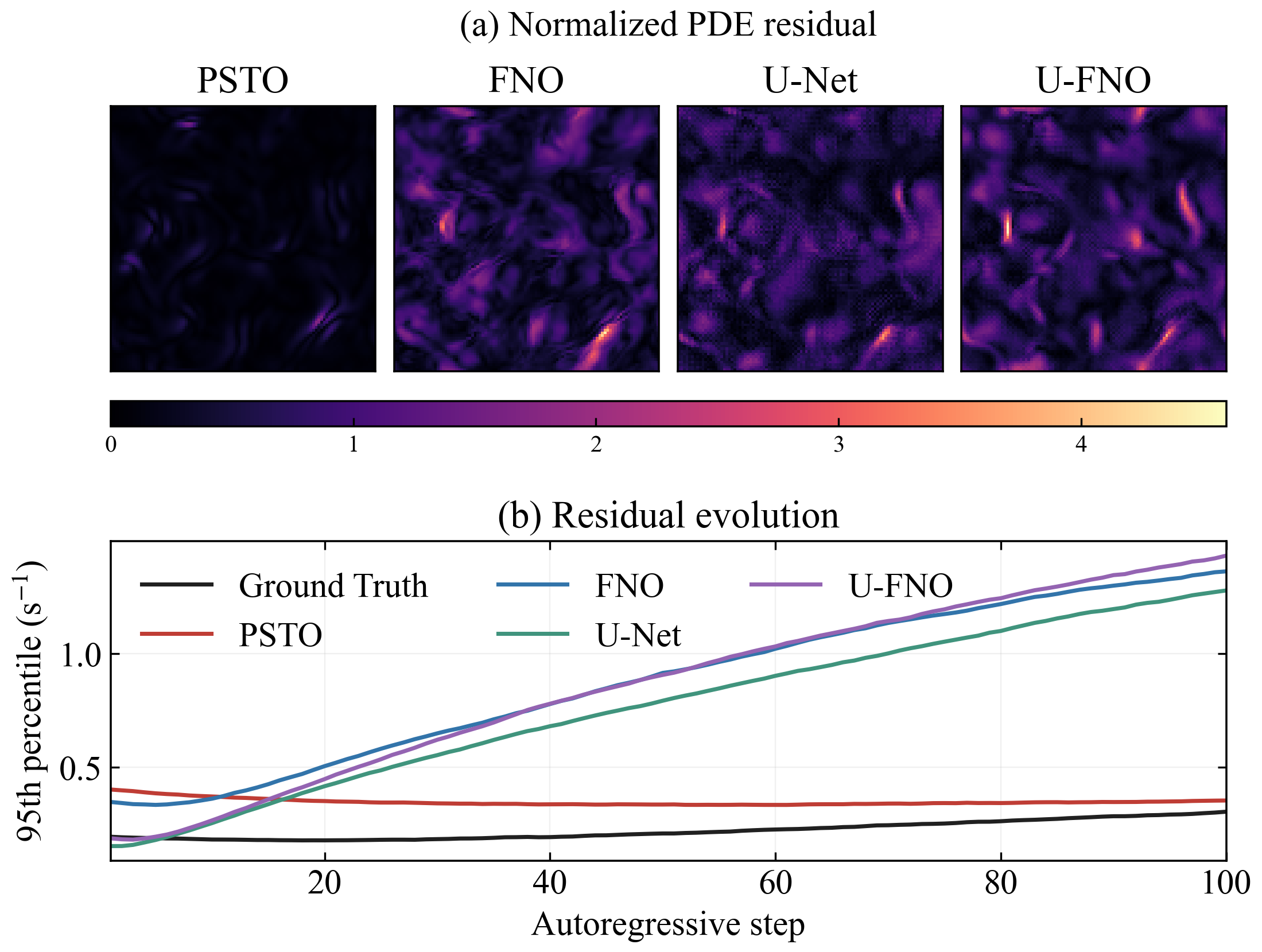}
		\caption{Canonical CDR residual defined by Eq.~\eqref{eq:canonical_rollout_residual}. (a) Endpoint residual fields on a common color scale. (b) Spatial 95th percentile over 100 coupled steps. The diagnostic includes the full governing equations of the canonical CDR system.}
		\label{fig:pde_residual_analysis}
	\end{figure}
	
	Figure~\ref{fig:pde_residual_analysis}(b) shows that the spatial 95th percentile of the PSTO residual remains between $0.335$ and $0.403\,\mathrm{s}^{-1}$ over the 100-step rollout. Its endpoint value of $0.354\,\mathrm{s}^{-1}$ is $12.0\%$ below its first-step value, whereas the baseline residuals reach $1.28$--$1.43\,\mathrm{s}^{-1}$, or 3.92--8.31 times their respective first-step values. The endpoint maps in Figure~\ref{fig:pde_residual_analysis}(a) show correspondingly lower residual magnitudes for PSTO on the same color scale. Together with the lower prediction errors in Table~\ref{tab:rel2_2d_CDR}, these results show that PSTO maintains close agreement with the reference trajectory while effectively suppressing physical deviations throughout the autoregressive rollout.
	
	\subsection{Two-dimensional Robertson counterflow}
	\label{sec:2d_robertson}
	
	The second case examines transport predictions under stiff reaction kinetics. The Robertson mechanism~\cite{robertson1966solution} comprises three reactions with widely separated kinetic time scales,
	\begin{equation}
		A\xrightarrow{k_1}B,\qquad B+B\xrightarrow{k_2}C+B,\qquad B+C\xrightarrow{k_3}A+C.
		\label{eq:robertson_reaction_scheme}
	\end{equation}
	The rate constants are $k_1=4\,\mathrm{s^{-1}}$, $k_2=3\times10^7\,\mathrm{m^3/(kmol \cdot s)}$, and $k_3=10^4\,\mathrm{m^3/(kmol \cdot s)}$. Species $A$, $B$, $C$, and inert $D$ share the molecular weight $W=25.9\,\mathrm{kg/kmol}$, and the mixture density is fixed at $1\,\mathrm{kg/(m^3)}$. Both the species diffusivity and kinematic viscosity are $10^{-4}\,\mathrm{m^2/s}$.
	
	The domain measures $0.02\,\mathrm{m}\times0.01\,\mathrm{m}$ and is discretized on a uniform $200\times100$ grid with $\Delta t=10^{-4}\,\mathrm{s}$. Opposing inlets prescribe streamwise velocities of approximately $+0.1$ and $-0.1\,\mathrm{m\,s^{-1}}$, while zero normal gradients are imposed at the transverse boundaries. The inlet composition is sampled from a randomly selected point in the reaction zone of a reacting counterflow. The species fields are then evolved with the reactions in Eq.~\eqref{eq:robertson_reaction_scheme} disabled to generate reaction-free CD training trajectories. The CDR reference data are generated by numerically solving the full counterflow equations with reactions enabled. The velocity fields remain steady in time, as the flow is unaffected by the reaction kinetics and all species share the same thermophysical properties. The predicted state therefore comprises only the four species mass fractions.
	
	For surrogate prediction, each learned transport map is coupled with the same adaptive SDIRK2 reaction solver using Strang composition. The CDR reference data are generated from the numerical simulations in OpenFOAM~\cite{jasak2007openfoam}. The training settings and reaction-solver details are given in Appendix~\ref{app:training_settings}. Field comparisons and spatial error maps are presented at step 100, corresponding to an elapsed time of $0.01\,\mathrm{s}$ or $10\%$ of the reference time $t_{\mathrm{ref}}=0.1\,\mathrm{s}$.
	
	Before comparing prediction accuracy, we examine how the sampled reacting states differ from the reaction-free states used for training the transport surrogates. Gaussian-kernel maximum mean discrepancy (MMD)~\cite{gretton2012kernel} quantifies this difference in a common principal-component representation, while a joint t-SNE embedding~\cite{van2008visualizing} visualizes the sample distributions. For this diagnostic, 100 chronologically ordered snapshots from each regime are selected approximately uniformly over frame indices 0--4000. Each snapshot is converted into a single feature vector by flattening and concatenating the six fields $(u,v,Y_A,Y_B,Y_C,Y_D)$. Each feature, corresponding to one variable at a specific grid point, is standardized using its mean and standard deviation across the pooled set of 200 snapshots from both regimes. PCA is then fitted to the standardized pooled data, and all snapshots are projected onto the leading 50 principal components to obtain a common low-dimensional representation. For the resulting samples $\mathcal X_{\rm CD}=\{\mathbf x_i\}_{i=1}^{M_{\rm CD}}$ and $\mathcal X_{\rm CDR}=\{\mathbf y_j\}_{j=1}^{M_{\rm CDR}}$, the empirical MMD is
	\begin{equation}
		\widehat{\mathrm{MMD}}=\left[\frac1{M_{\rm CD}^2}\sum_{i,i'}k(\mathbf x_i,\mathbf x_{i'})+\frac1{M_{\rm CDR}^2}\sum_{j,j'}k(\mathbf y_j,\mathbf y_{j'})-\frac2{M_{\rm CD}M_{\rm CDR}}\sum_{i,j}k(\mathbf x_i,\mathbf y_j)\right]^{1/2}, k(\mathbf x,\mathbf y)=e^{-\|\mathbf x-\mathbf y\|_2^2/(2\sigma_{\rm ker}^2)},
		\label{eq:robertson_mmd}
	\end{equation}
	where $\sigma_{\rm ker}$ is the median pairwise distance in the pooled PCA representation. For each within-regime comparison, the 100 snapshots are divided into two interleaved groups of 50 using alternating sample positions. All within-regime and cross-regime comparisons use the same standardization, PCA transformation, and kernel bandwidth.
	
	\begin{figure}[pos=htbp]
		\centering
		\includegraphics[width=0.6\linewidth]{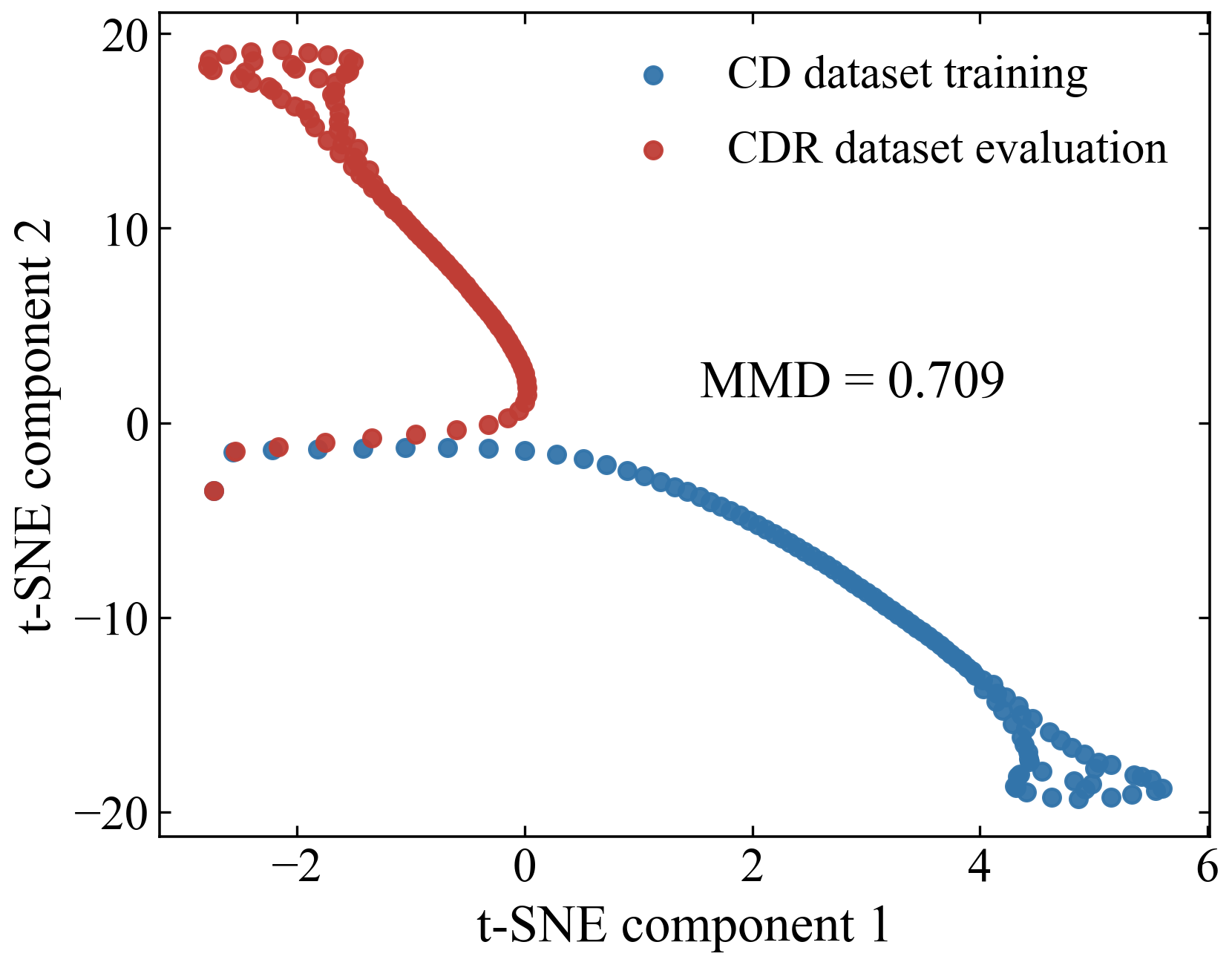}
		\caption{Distribution shift between Robertson CD training states (blue) and CDR evaluation states (red), visualized through a joint t-SNE embedding of 100 snapshots per regime. The cross-regime MMD, computed in the pooled PCA representation, is 0.709.}
		\label{fig:robertson_inference_transfer}
	\end{figure}
	
\begin{figure}[pos=htbp]
		\centering
		\includegraphics[width=0.98\linewidth]{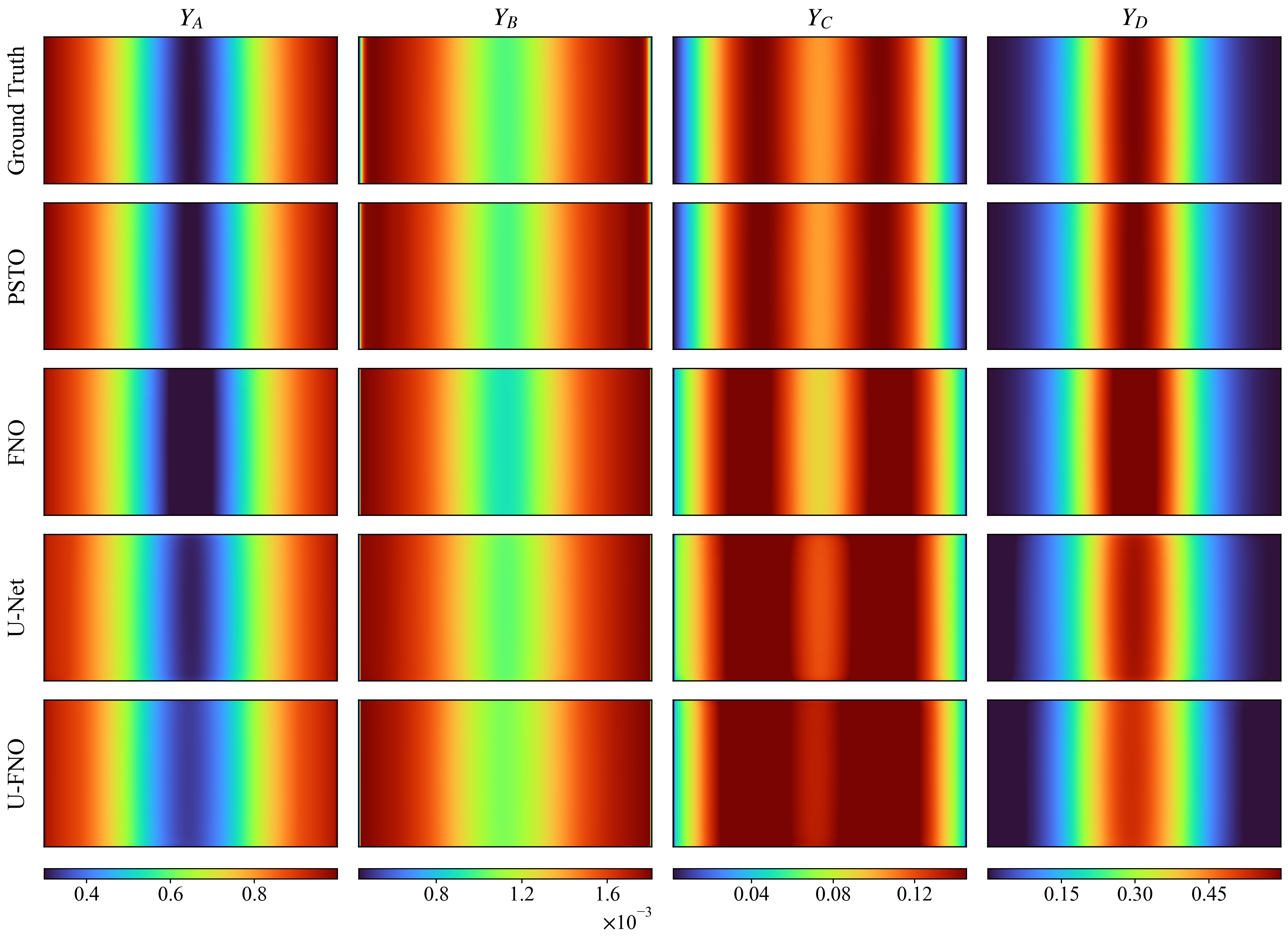}
		\caption{Robertson counterflow mass fractions after 100 coupled steps ($t-t_0=0.01\,\mathrm{s}$). Columns show $Y_A,Y_B,Y_C,Y_D$, and rows show the numerical reference, PSTO, FNO, U-Net, and U-FNO. All models use the same velocity field, inlet data, and reaction kinetics.}
		\label{fig:2d_Robert100_results}
	\end{figure}

	The cross-regime MMD is $0.709$, compared with within-regime values of $0.0260$ for CD and $0.0262$ for CDR. Figure~\ref{fig:robertson_inference_transfer} further illustrates this pronounced distribution shift between the training and evaluation datasets. Despite this shift, PSTO maintains accurate out-of-distribution predictions, as shown in Table~\ref{tab:rel2_2d_robert} and Fig.~\ref{fig:2d_Robert100_results}, supporting the generalizability of its PDE-structured design.
	
	\begin{table}[pos=htbp]
		\centering
		\caption{Mean variable-wise relative $L_2$ error (\%) for Robertson counterflow, evaluated using Eq.~\eqref{eq:evaluation_metric}. Columns indicate the number of coupled steps.}
		\label{tab:rel2_2d_robert}
		\small
		\setlength{\tabcolsep}{4pt}
		\begin{tabular}{lrrrrrr}
			\toprule
			Model & 1 & 5 & 10 & 20 & 50 & 100 \\
			\midrule
			PSTO & 0.364 & 0.914 & 1.04 & 1.13 & 1.30 & 1.50 \\
			FNO & 0.640 & 2.05 & 2.75 & 3.84 & 7.09 & 12.4 \\
			U-Net & 0.641 & 2.04 & 2.71 & 3.72 & 6.51 & 10.6 \\
			U-FNO & 0.662 & 2.14 & 2.93 & 4.16 & 7.61 & 12.6 \\
			\bottomrule
		\end{tabular}
	\end{table}
	
	Table~\ref{tab:rel2_2d_robert} shows that PSTO achieves the lowest error at every reported step. Between steps 20 and 100, its error increases from $1.13\%$ to $1.50\%$, whereas the error of U-Net, the most accurate baseline, rises from $3.72\%$ to $10.6\%$. PSTO thus reduces the endpoint error by approximately $85.8\%$ relative to U-Net. Figure~\ref{fig:2d_Robert100_results} shows that PSTO more accurately resolves the two regions of elevated $Y_C$ and the central minimum separating them, whereas the baselines produce more diffuse product distributions. The lower error in the inert species $Y_D$ shows that the improvement also extends to a species with no reaction source. Since all models use the same reaction solver, the comparison evaluates how differences in transport predictions affect the subsequent coupled evolution, as described by Eq.~\eqref{eq:complete_error_bound}.
	
	\subsection{Two-dimensional non-premixed planar jet flame}
	\label{sec:2d_planarjet}
	
	The final case evaluates PSTO in a twelve-field, multiscale CDR problem with spatially varying thermophysical properties and learned local reaction updates. The hydrogen planar jet flame is simulated in a $0.05\,\mathrm{m}\times0.03\,\mathrm{m}$ domain, discretized on a uniform $500\times300$ grid with $\Delta x=10^{-4}\,\mathrm{m}$ and $\Delta t=10^{-6}\,\mathrm{s}$. The predicted state comprises temperature $T$, nine species mass fractions $(\mathrm{H_2},\mathrm H,\mathrm O,\mathrm{O_2},\mathrm{OH},\mathrm{H_2O},\mathrm{HO_2},\mathrm{H_2O_2},\mathrm{N_2})$, and the velocity components $(u,v)$. The hydrogen mechanism is evaluated using Cantera~\cite{goodwin2018cantera} to prepare the inlet composition for CD trajectories and the numerical references. The inlet configuration, initial and boundary conditions are detailed in Appendix~\ref{app:planarjet_setting}.
	
	Training data for the transport surrogates are generated from reaction-free CD simulations. The central fuel-stream composition is replaced by an equilibrium composition computed with a zero-dimensional reactor in Cantera, while the adjacent streams supply air. These fields evolve through convection and diffusion with chemical source terms disabled. The CDR reference simulations use OpenFOAM with chemical reactions enabled. Nitrogen-diluted hydrogen enters through the central inlet, and air enters on both sides. Coupled prediction starts from a developed reacting state.
	
	The local reaction model comprises eight networks that predict reactive-species increments. Each network has four hidden layers of width 400 and takes $(T,p,f_{\rm BCT}(\mathbf Y))$ as input. The Box--Cox exponent is $\beta=0.1$. The reaction update retains zero direct increment for inert nitrogen. The resulting species vector is then normalized to unit sum and temperature is recovered from mixture enthalpy conservation. The constitutive network supplies $(\rho,\mu,\lambda/c_p)$ from the current thermochemical state, and species diffusivities are obtained using the corresponding Lewis numbers. The spatially varying transport coefficients are retained inside the weighted flux divergence, accounting for nonuniform transport properties across the fuel--air mixing layers and flame regions.
	
	For surrogate prediction, the learned transport and reaction maps are coupled using Lie--Trotter splitting, with a full transport step followed by a full reaction step. Field comparisons and spatial error maps are presented at step 100, corresponding to an elapsed time of $10^{-4}\,\mathrm{s}$ or $13.1\%$ of the reference time $t_{\mathrm{ref}}=7.64\times10^{-4}\,\mathrm{s}$.
	
	\begin{table}[pos=htbp]
		\centering
		\caption{Mean variable-wise relative $L_2$ error (\%) for the hydrogen planar jet flame, evaluated using Eq.~\eqref{eq:evaluation_metric}. Columns indicate the number of coupled steps.}
		\label{tab:rel2_2d_planarjet}
		\small
		\setlength{\tabcolsep}{4pt}
		\begin{tabular}{lrrrrrr}
			\toprule
			Model & 1 & 5 & 10 & 20 & 50 & 100 \\
			\midrule
			PSTO & 1.03 & 3.55 & 5.79 & 9.28 & 21.4 & 23.3 \\
			FNO & 1.45 & 6.35 & 12.7 & 23.5 & 44.4 & 51.4 \\
			U-Net & 1.74 & 7.57 & 12.6 & 20.9 & 40.9 & 50.5 \\
			U-FNO & 1.49 & 7.26 & 12.6 & 21.2 & 41.0 & 48.5 \\
			\bottomrule
		\end{tabular}
	\end{table}
	
	Table~\ref{tab:rel2_2d_planarjet} shows that PSTO achieves the lowest error at every reported horizon. After 100 steps, its mean error across the twelve variables is $23.3\%$, compared with $48.5$--$51.4\%$ for the baselines. This corresponds to an error reduction of approximately $51.9\%$ relative to the best-performing baseline, U-FNO.
	
	\begin{figure}[pos=htbp]
		\centering
		\includegraphics[width=0.98\linewidth]{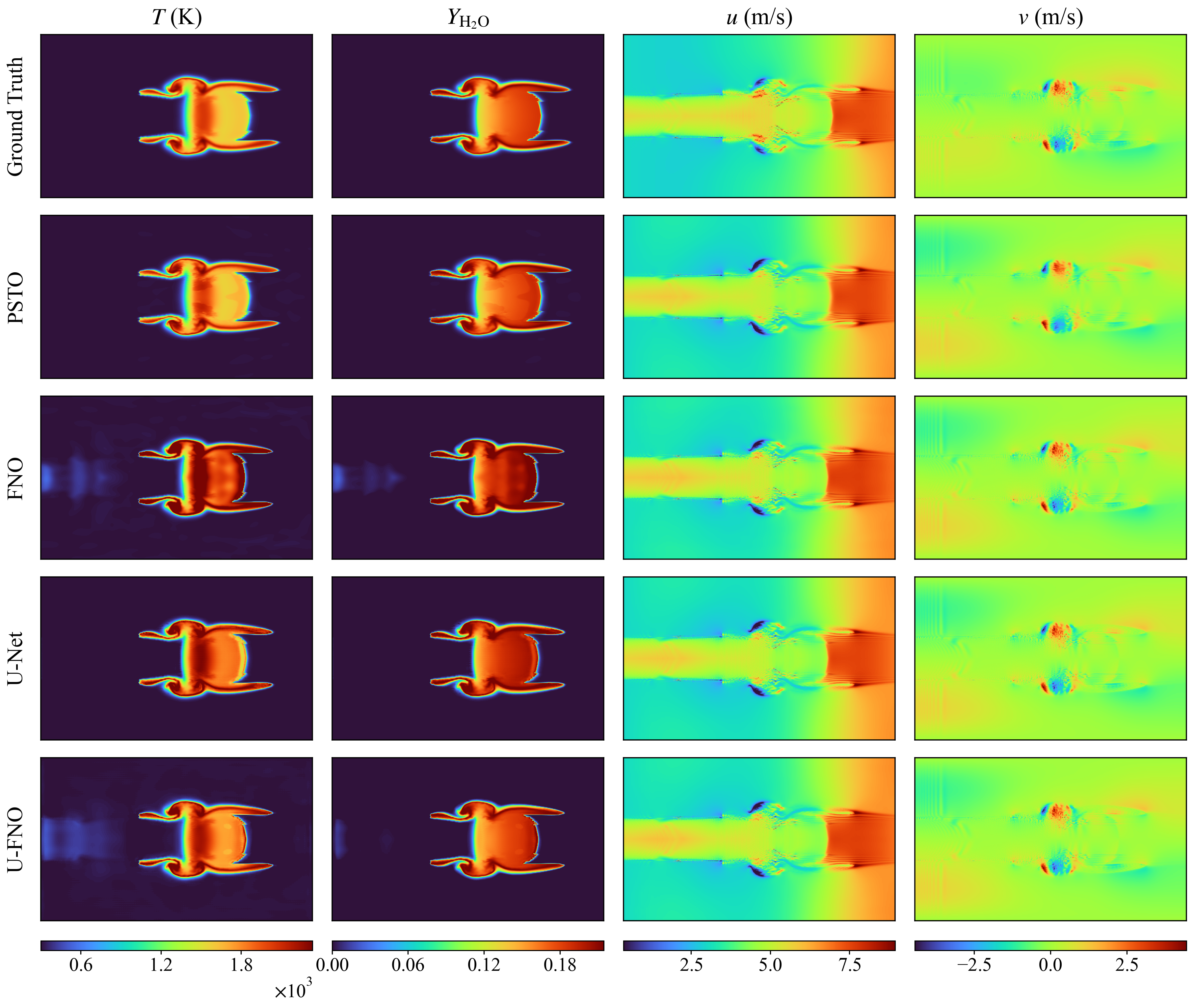}
		\caption{Planar jet flame fields after 100 coupled steps ($t-t_0=10^{-4}\,\mathrm{s}$). Columns show $T$, $Y_{\mathrm{H_2O}}$, $u$, and $v$, and rows show the numerical reference, PSTO, FNO, U-Net, and U-FNO. All surrogate models use the same initial state and Lie--Trotter transport--reaction coupling.}
		\label{fig:2d_PlanarJet100_results}
	\end{figure}
	
	Figure~\ref{fig:2d_PlanarJet100_results} shows that PSTO more accurately captures the downstream high-temperature region and curved flame structures. Its temperature and water-mass-fraction errors are $5.28\%$ and $6.57\%$, respectively, corresponding to reductions of $72.1\%$ and $72.5\%$ relative to the best baseline for each variable. The individual errors are reported in Table~\ref{tab:channels_PlanarJet}. The baselines exhibit larger deviations in these fields, and FNO and U-FNO also produce visible upstream artifacts. PSTO achieves the lowest relative error among the evaluated models for each of the twelve variables. Nevertheless, the relative errors in the five minor species and the transverse velocity remain comparatively large, as their small reference magnitudes increase sensitivity to absolute prediction errors. These contributions raise the mean relative error across all twelve variables above the individual errors in temperature and water mass fraction. Appendix~\ref{app:spatial_errors} shows the spatial distribution of the temperature error.
	
	\subsection{Prediction accuracy, stability, and efficiency}
    \label{sec:error_propagation}
    
    The preceding cases compare the accuracy of autoregressive coupled predictions. To assess PSTO more comprehensively, we now examine the effect of latent iteration depth, quantify how one-step coupled errors propagate during rollouts, and evaluate stability and inference efficiency alongside prediction accuracy.
	
	\subsubsection{Fixed-point residual decay and prediction accuracy}
	\label{sec:iteration_ablation}
	
	To isolate the effect of iteration depth, we evaluate a PSTO model trained with four shared iterations using only its first one, two, three, or four iterations. The test uses 64 uniformly spaced Robertson CD validation pairs. For each pair, all trained parameters, including the lifting and projection maps, are held fixed. The input state and conditioning fields also remain unchanged, so prediction differences arise solely from the number of latent updates.
	
\begin{figure}[pos=htbp]
        \centering
        \includegraphics[width=0.98\linewidth]{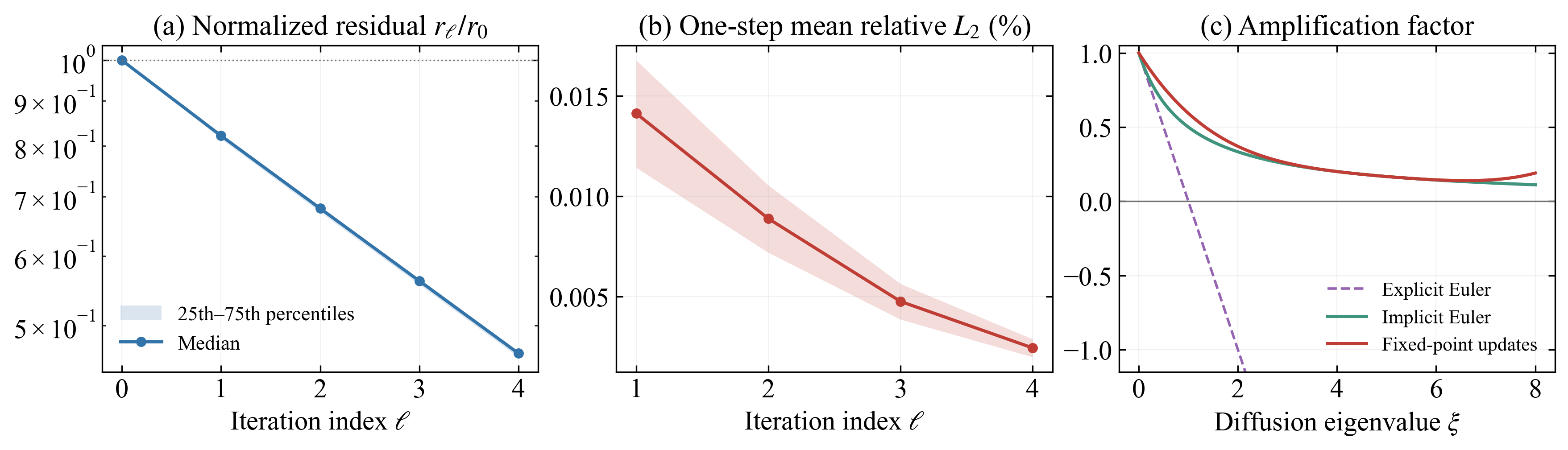}
        \caption{Iterative refinement and illustrative diffusion-mode analysis. (a) Normalized fixed-point residual $r_\ell/r_0$ evaluated at $z^\ell$ and (b) one-step relative error averaged over variables, evaluated on 64 Robertson CD validation pairs with trained weights held fixed. Lines indicate medians and bands represent interquartile ranges. (c) Amplification factors for explicit Euler, implicit Euler, and four relaxed fixed-point updates at $\omega=0.1718$, plotted against the diffusion eigenvalue $\xi=\Delta t\lambda_{\rm mode}$, where $\lambda_{\rm mode}$ is the modal decay rate and $\Delta t$ is the time step.}
        \label{fig:iteration_ablation}
    \end{figure}

    Figures~\ref{fig:iteration_ablation}(a,b) show that additional relaxed fixed-point iterations reduce both the normalized implicit residual and the one-step transport prediction error at fixed trained weights. Increasing the iteration count from one to four reduces the median one-step relative error averaged over variables from $0.0141\%$ to $0.00245\%$, a 5.8-fold reduction, while narrowing its interquartile range. These trends are consistent with Proposition~\ref{prop:iteration_fixed_point}, where reducing the iteration residual defined in Eq.~\eqref{eq:reference_residual_and_projection_error} tightens the error bound in Eq.~\eqref{eq:projected_implicit_accuracy}.
    
    To illustrate the stability mechanism, consider an idealized scalar diffusion mode governed by $\partial_t z=\mathcal T(z)=-\lambda_{\rm mode}z$, where $\lambda_{\rm mode}\geq0$ is the decay rate. Define the scaled diffusion eigenvalue $\xi=\Delta t\lambda_{\rm mode}$ associated with the operator $-\Delta t\mathcal T$. The amplification factors for implicit Euler, explicit Euler, and $L$ relaxed fixed-point updates starting from the input state are
    \begin{equation}
        g_{\rm i}(\xi)=\frac{1}{1+\xi},\qquad g_{\rm e}(\xi)=1-\xi,\qquad g_L(\xi)=g_{\rm i}(\xi)+\frac{\xi}{1+\xi}[1-\omega(1+\xi)]^L.
        \label{eq:modal_amplification}
    \end{equation}
    Explicit Euler produces sign reversal for $\xi>1$ and amplitude growth for $\xi>2$. The relaxed fixed-point iteration converges to implicit Euler throughout $0\leq\xi\leq\xi_{\max}$ provided that $0<\omega<2/(1+\xi_{\max})$.
    
    Figure~\ref{fig:iteration_ablation}(c) compares these amplification factors using the learned relaxation parameter $\omega=0.1718$ from Figs.~\ref{fig:iteration_ablation}(a,b). Four relaxed fixed-point updates yield $0<g_{L=4}(\xi)\leq1$ over $0\leq\xi\leq8$, including modes for which explicit Euler exhibits sign reversal or amplitude growth. This idealized example illustrates the potential stability benefits of finite relaxed fixed-point updates over explicit time stepping. For the full learned convection--diffusion operator, convergence is established under the joint conditions in Eq.~\eqref{eq:full_transport_conditions}.
    
	\subsubsection{Error decomposition, local transport sensitivity, and autoregressive propagation}
	
	We next examine how PSTO's one-step transport error and error amplification relate to the coupled one-step error $\delta_{\Delta t}^{\chi}$ and autoregressive prediction error $e_n$ in Eq.~\eqref{eq:split_rollout_error}. To this end, we analyze error decomposition, local sensitivity, and autoregressive error propagation along another planar jet flame trajectory. The trajectory comprises an initial state and 100 successive reference states, defining a 100-step rollout. OpenFOAM with chemistry disabled provides the numerical transport map $\mathcal T^{\rm ref}$, while Cantera provides the numerical reaction map $\mathcal R^{\rm ref}$ over the same time interval $\Delta t$. For each model, $\widehat{\mathcal T}$ denotes the implemented transport map. All transport models share the reaction surrogate $\widehat{\mathcal R}$, and their composition defines the coupled Lie--Trotter map $\widehat{\mathcal S}=\widehat{\mathcal R}\circ\widehat{\mathcal T}$. 
	
	For the sensitivity analysis, perturbed states $\widetilde q_n$ are generated by adding spatially smoothed perturbations with magnitudes of $10^{-4}$ relative to each variable's range, while preserving boundary values and species admissibility. The one-step relative $L_2$ error and local transport sensitivity are defined as
    \begin{equation}
        \delta_n= \frac{\|\widehat{\mathcal S}(q_n)-q_{n+1}\|_2} {\|q_{n+1}\|_2}, \qquad \Lambda_n^{\rm loc}= \frac{\|\widehat{\mathcal T}(\widetilde q_n) -\widehat{\mathcal T}(q_n)\|_2} {\|\widetilde q_n-q_n\|_2}.
        \label{eq:sampled_coupled_diagnostics}
    \end{equation}
    Here, $\widehat q_n$ denotes the autoregressively predicted state, and $q_n$ denotes the numerical reference state. The normalized autoregressive error $e_n$ is evaluated using the definition in Proposition~\ref{prop:coupled_error_bound}, applied here to the predicted and numerical reference trajectories $(\widehat q_n,q_n)$ with initial state $q_0$. All reference norms in the denominators are assumed nonzero.
	
	We decompose the one-step prediction error into three contributions. The transport contribution measures the change in the coupled prediction when learned transport replaces numerical transport. The reaction surrogate remains fixed in this comparison. The reaction contribution compares the common reaction surrogate with the reference reaction map, both applied to the numerical transport output. The remaining contribution compares the composed transport--reaction reference with the next state of the full CDR reference trajectory.
	\begin{equation}
			d_n^{\rm T} =\widehat{\mathcal R}(\widehat{\mathcal T}(q_n)) -\widehat{\mathcal R}(\mathcal T^{\rm ref}(q_n)),\qquad d_n^{\rm R} =\widehat{\mathcal R}(\mathcal T^{\rm ref}(q_n)) -\mathcal R^{\rm ref}(\mathcal T^{\rm ref}(q_n)),\qquad d_n^{\rm S} =\mathcal R^{\rm ref}(\mathcal T^{\rm ref}(q_n))-q_{n+1}.
	\end{equation}
	By construction, these three vectors sum exactly to the complete one-step prediction error. The reference trajectory containing $q_{n+1}$ is also generated using operator splitting in OpenFOAM, with an integration time step substantially smaller than the prediction interval to reduce temporal discretization error. Applying the triangle inequality yields
	\begin{equation}
		\delta_n =\frac{\|d_n^{\rm T}+d_n^{\rm R}+d_n^{\rm S}\|_2}{\|q_{n+1}\|_2} \leq \frac{\|d_n^{\rm T}\|_2+\|d_n^{\rm R}\|_2+\|d_n^{\rm S}\|_2}{\|q_{n+1}\|_2}.
		\label{eq:sampled_one_step_bound}
	\end{equation}
	
\begin{figure}[pos=htbp]
		\centering
		\includegraphics[width=0.98\linewidth]{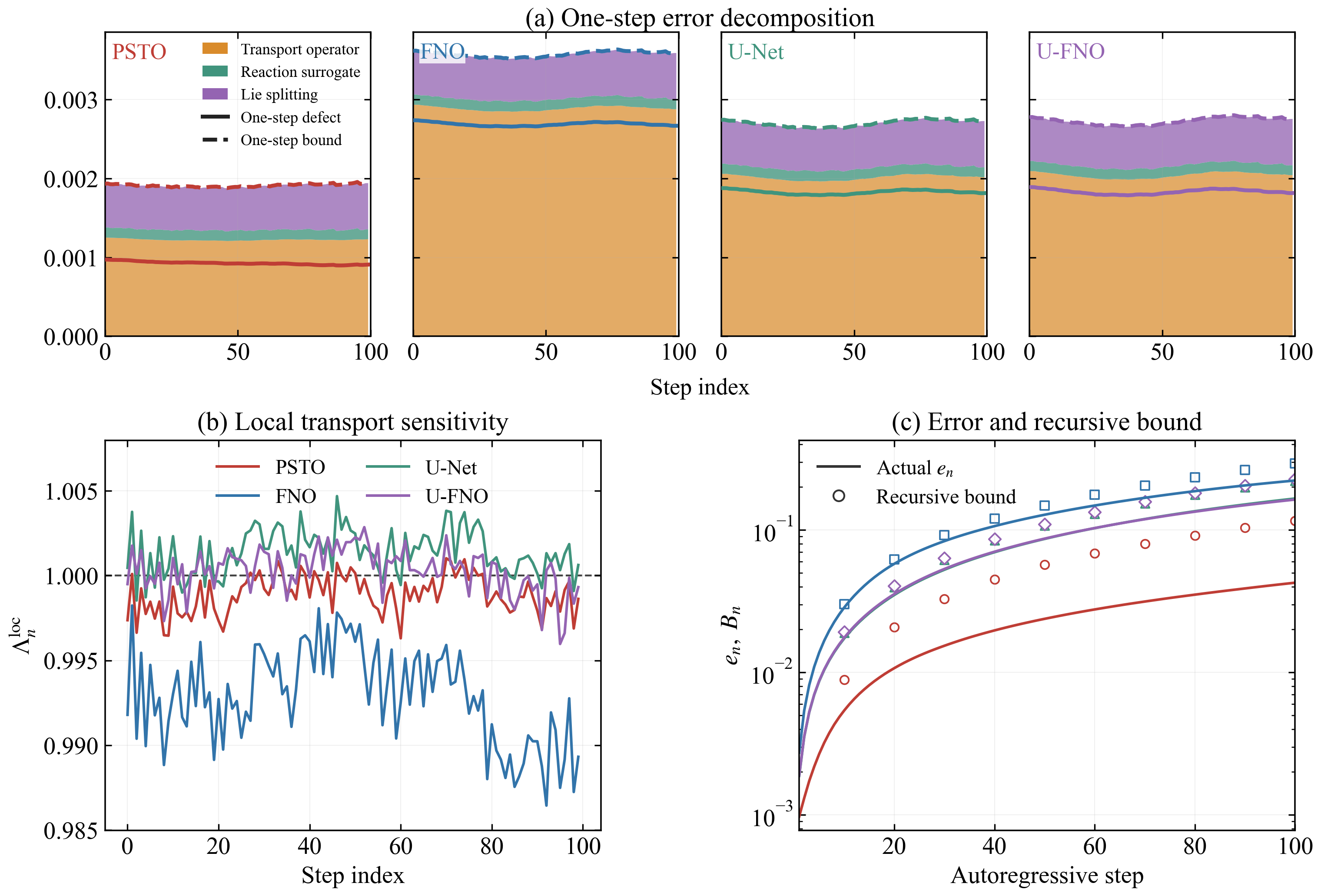}
		\caption{One-step error decomposition, transport sensitivity, and autoregressive error propagation for the planar jet flame. (a) Transport, reaction, and splitting contributions to the one-step error. The measured errors and summed contribution norms are shown for comparison. (b) Local transport sensitivity to small perturbations at successive reference states. The dashed line marks $\Lambda_n^{\mathrm{loc}}=1$, indicating unchanged perturbation magnitude after one transport update. (c) Relative autoregressive errors $e_n$ and recursive bounds $B_n$.}
		\label{fig:coupled_sensitivity_error}
	\end{figure}

	Figure~\ref{fig:coupled_sensitivity_error}(a) compares the measured one-step errors with the three error contributions. PSTO has the smallest relative transport contribution of $1.28\times10^{-3}$, compared with $(1.96\text{--}2.76)\times10^{-3}$ for the baselines. The contribution from the shared reaction surrogate is approximately one order of magnitude smaller. PSTO also achieves the smallest measured one-step error $9.28\times10^{-4}$, compared with $(1.83\text{--}2.69)\times10^{-3}$ for the baselines. The summed contributions consistently bound the measured errors, in agreement with Eq.~\eqref{eq:sampled_one_step_bound}. The gap reflects the use of the triangle inequality, which sums the norms of the individual error components without accounting for their possible cancellation.
	
    Figure~\ref{fig:coupled_sensitivity_error}(b) provides numerical evidence of PSTO's local stability under the tested perturbations. Across 100 consecutive reference states, its local transport sensitivity remains close to unity, with a mean of $0.9989$. Thus, a single transport update does not substantially amplify these perturbations. This response is relevant to autoregressive prediction because accumulated errors perturb the inputs to subsequent updates. Equation~\eqref{eq:coupled_lipschitz_constant} links transport sensitivity to coupled error amplification. With the reaction surrogate fixed, a smaller uniform transport bound tightens the upper bound on coupled amplification. This coupled bound enters successive powers in Eq.~\eqref{eq:split_rollout_error}, so controlling transport sensitivity can limit the accumulation of one-step errors over longer rollouts. Together with the small one-step errors in Fig.~\ref{fig:coupled_sensitivity_error}(a), these results support PSTO's potential for accurate and stable coupled prediction. We next examine error propagation along the actual coupled autoregressive trajectory.
		
	We evaluate the complete coupled map at both the predicted and reference states. Using $\delta_n$ from Eq.~\eqref{eq:sampled_coupled_diagnostics} and expressing the one-step contribution relative to the initial-state norm, we construct the recursive bound
    \begin{equation}
        B_1 = \frac{\|q_1\|_2}{\|q_0\|_2}\delta_0, \qquad B_{n+1}= \frac{\|q_{n+1}\|_2}{\|q_0\|_2}\delta_n +\alpha_n B_n, \quad n\geq1, \qquad \alpha_n = \frac{\|\widehat{\mathcal S}(\widehat q_n)-\widehat{\mathcal S}(q_n)\|_2}{\|\widehat q_n-q_n\|_2}.
        \label{eq:sampled_recursive_bound}
    \end{equation}
	Here, $\alpha_n$ measures how the complete coupled update amplifies the prediction error at step $n$. We set $\alpha_n=0$ when $\widehat q_n=q_n$, since the numerator also vanishes. For a shared initial state, the recurrence gives $e_n\leq B_n$ along the evaluated trajectory, with both quantities normalized by $\|q_0\|_2$. Figure~\ref{fig:coupled_sensitivity_error}(c) shows that PSTO maintains the lowest prediction error and recursive bound throughout the 100-step rollout. Its final relative error is $4.26\times10^{-2}$, approximately $74$--$81\%$ lower than those of the baselines ($0.163$--$0.222$). FNO accumulates the largest prediction error despite its stronger attenuation of the sampled transport perturbations. Together, these results show that one-step accuracy and error amplification by the complete coupled update jointly govern autoregressive prediction errors.
	
	\subsubsection{Inference time and prediction accuracy}
	\label{sec:inference_cost}
	
	To quantify the computational cost associated with PSTO's accuracy gains, we compare its coupled rollout time with those of the neural baselines and the numerical reference workflow. For the planar jet flame benchmark, each neural model advances the same randomly selected reacting initial state through 100 coupled transport--reaction steps at $\Delta t=10^{-6}\,\mathrm{s}$, corresponding to a physical horizon of $10^{-4}\,\mathrm{s}$. All four neural models are evaluated on a single NVIDIA GeForce RTX 4090 GPU. Two warmup rollouts precede five timed repetitions, with GPU synchronization at the beginning and end of each measured rollout. The measured workload includes the constitutive MLP where applicable, transport, the reaction MLPs, enforcement of species constraints, and enthalpy-conserving temperature recovery.
	
	\begin{figure}[pos=htbp]
		\centering
		\includegraphics[width=0.98\linewidth]{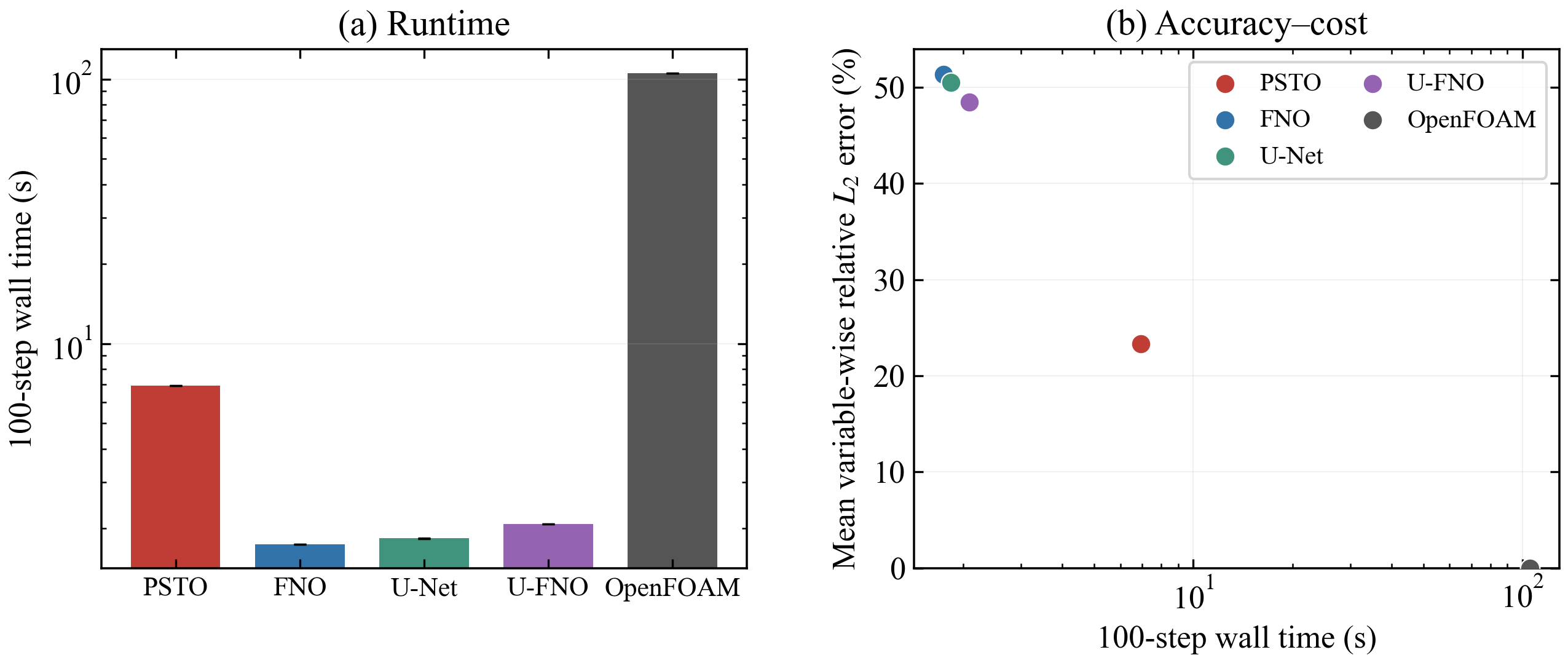}
		\caption{Computational performance for the planar jet flame over a physical horizon of $10^{-4}\,\mathrm{s}$, corresponding to 100 coupled steps at $\Delta t=10^{-6}\,\mathrm{s}$ for the neural models. (a) Coupled inference wall time averaged over five timed repetitions for the neural models, together with the runtime of the single OpenFOAM run. (b) Mean variable-wise relative $L_2$ error at the common endpoint versus wall time. The neural models use a single RTX 4090 GPU, whereas OpenFOAM uses 128 CPU MPI ranks. OpenFOAM serves as the numerical reference and is therefore shown at zero error by definition.}
		\label{fig:computational_performance}
	\end{figure}
	
	As shown in Fig.~\ref{fig:computational_performance}(a), PSTO requires $6.94\,\mathrm{s}$, compared with $1.74$--$2.08\,\mathrm{s}$ for the neural baselines. Its higher inference cost reflects the repeated iterations within the transport block. The OpenFOAM workflow requires $105.42\,\mathrm{s}$ on 128 CPU MPI ranks, giving PSTO an approximately 15-fold wall-time speedup under the stated hardware configurations. Figure~\ref{fig:computational_performance}(b) illustrates the corresponding accuracy--cost trade-off: PSTO reduces the mean endpoint error by approximately $51.9\%$ relative to U-FNO, at approximately 3.3 times the inference wall time.
	
	\subsubsection{Time-step sensitivity and prediction stability}
	\label{sec:timestep_stability}
	
	To assess time-step sensitivity and prediction stability, we repeat the planar jet flame evaluation with a tenfold larger surrogate time step, increasing $\Delta t$ from $10^{-6}$ to $10^{-5}\,\mathrm{s}$. For this comparison, all four transport surrogates and the common local reaction surrogate are trained for the larger time interval. Each transport surrogate is coupled to the same reaction surrogate, and the initial state, coupling procedure, and evaluation protocol remain unchanged. 10 coupled steps at $\Delta t=10^{-5}\,\mathrm{s}$ span $10^{-4}\,\mathrm{s}$, matching the physical horizon of the 100-step predictions at $\Delta t=10^{-6}\,\mathrm{s}$. Table~\ref{tab:rel2_2d_planarjet_dt1e5} reports the mean variable-wise relative $L_2$ errors at selected prediction horizons.
	
	\begin{table}[pos=htbp]
		\centering
		\caption{Mean variable-wise relative $L_2$ errors (\%) for the hydrogen planar jet flame at $\Delta t=10^{-5}\,\mathrm{s}$, evaluated using Eq.~\eqref{eq:evaluation_metric}. Column headings indicate the number of coupled steps. The corresponding physical horizons match those of steps 10, 20, 50, and 100 at $\Delta t=10^{-6}\,\mathrm{s}$ in Table~\ref{tab:rel2_2d_planarjet}. All four transport surrogates are coupled to the same reaction surrogate trained for the larger time interval.}
		\label{tab:rel2_2d_planarjet_dt1e5}
		\small
		\setlength{\tabcolsep}{4pt}
		\begin{tabular}{lrrrr}
			\toprule
			Model & 1 & 2 & 5 & 10 \\
			\midrule
			PSTO & 5.96 & 9.16 & 21.0 & 23.4 \\
			FNO & 19.5 & 33.7 & 68.7 & 74.7 \\
			U-Net & 13.0 & 23.7 & 46.5 & 57.2 \\
			U-FNO & 13.3 & 23.9 & 48.0 & 59.0 \\
			\bottomrule
		\end{tabular}
	\end{table}
	
	PSTO's mean endpoint error remains nearly unchanged, increasing from $23.3\%$ to $23.4\%$, whereas the baseline errors increase to $57.2$--$74.7\%$. PSTO also achieves the lowest mean error at every reported horizon and the lowest error for each of the twelve variables at the endpoint, as detailed in Appendix Table~\ref{tab:channels_PlanarJet_dt1e5}. These results indicate that PSTO maintains comparable aggregate prediction accuracy at a tenfold larger time step over the tested physical horizon, while the baselines exhibit greater sensitivity to the time step.
	
\begin{figure}[pos=htbp]
		\centering
		\includegraphics[width=0.98\linewidth]{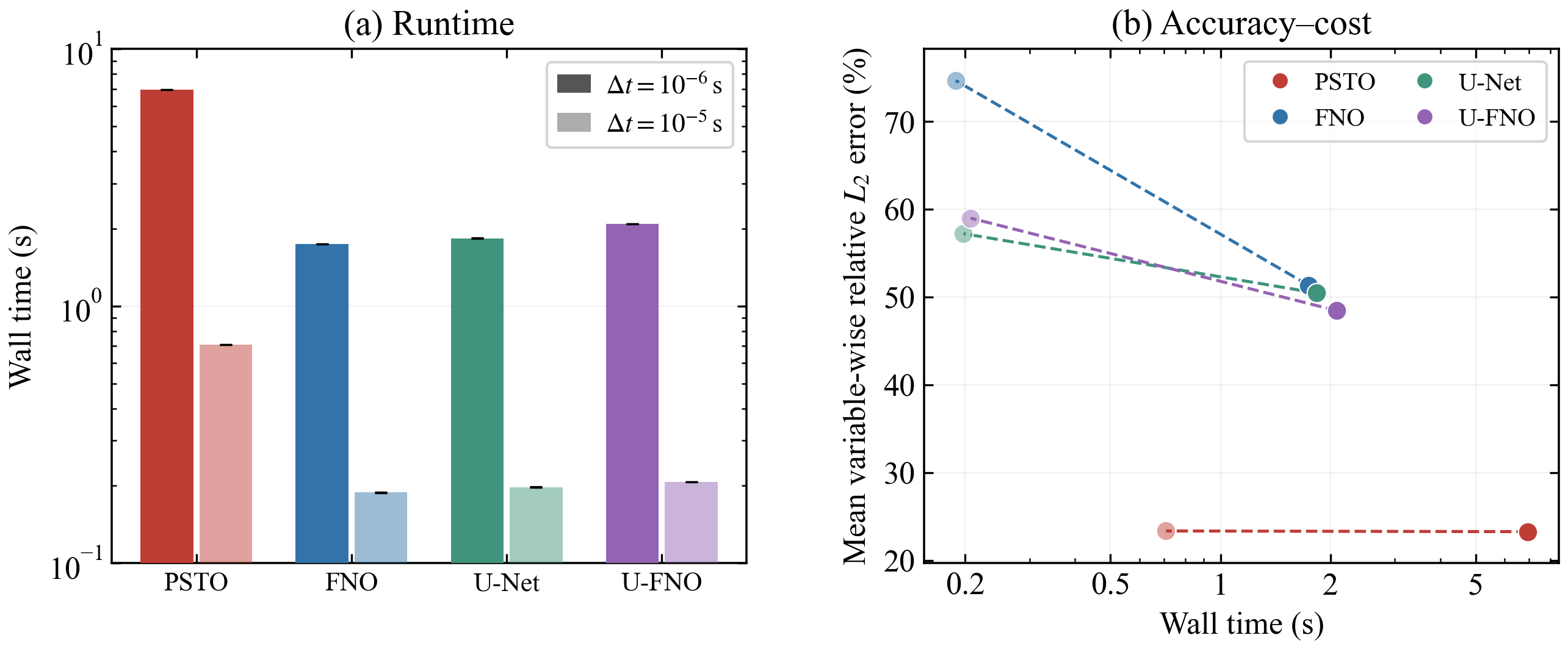}
		\caption{Time-step dependence of coupled planar jet flame predictions over a common physical horizon of $10^{-4}\,\mathrm{s}$, starting from the same initial state. (a) Coupled inference wall time at $\Delta t=10^{-6}\,\mathrm{s}$ (left, opaque) and $\Delta t=10^{-5}\,\mathrm{s}$ (right, translucent) within each model pair. Bars show mean wall times over five timed repetitions on a single RTX 4090 GPU. (b) Mean variable-wise relative $L_2$ error at the common endpoint versus wall time. Colors identify the models, opacity distinguishes the time steps, and dashed lines connect the corresponding pairs.}
		\label{fig:computational_performance_timestep}
	\end{figure}

	Table~\ref{tab:runtime_planarjet_timesteps} and Fig.~\ref{fig:computational_performance_timestep}(a) show that increasing the time step reduces PSTO's coupled inference wall time from $6.94$ to $0.709\,\mathrm{s}$. This represents an approximately 9.8-fold reduction over the same physical horizon, with essentially unchanged mean endpoint error. At the larger time step, PSTO achieves an approximately 149-fold wall-time speedup over the OpenFOAM workflow. This comparison uses the hardware configurations reported in Section~\ref{sec:inference_cost}, where OpenFOAM requires $105.42\,\mathrm{s}$ on 128 CPU MPI ranks.

    Although the baselines remain faster when all models are evaluated at $\Delta t=10^{-5}\,\mathrm{s}$, their prediction errors increase substantially. Figure~\ref{fig:computational_performance_timestep}(b) highlights PSTO's favorable accuracy--cost trade-off: at $\Delta t=10^{-5}\,\mathrm{s}$, PSTO is both faster and more accurate than all three baselines at $\Delta t=10^{-6}\,\mathrm{s}$. Its higher per-step cost can therefore be offset by using a larger time step while retaining comparable mean endpoint accuracy. These observations are consistent with the intended benefits of PSTO's implicit-like design and support its prediction robustness over the tested time steps and physical horizon.

    \begin{table}[pos=htbp]
		\centering
		\caption{Mean coupled inference wall time (s) for the hydrogen planar jet flame over a common physical horizon of $10^{-4}\,\mathrm{s}$. The two time steps require 100 and 10 coupled steps, respectively. Measurements follow the protocol in Section~\ref{sec:inference_cost}, using a single RTX 4090 GPU, two warmup rollouts, and five timed repetitions.}
		\label{tab:runtime_planarjet_timesteps}
		\small
		\setlength{\tabcolsep}{4pt}
		\begin{tabular}{lrrrr}
			\toprule
			\raisebox{-0.45\height}{%
				\setlength{\unitlength}{1pt}%
				\begin{picture}(32,20)
					\put(0,20){\rotatebox{-32.005}{\rule{37.736pt}{0.4pt}}}
					\put(31,17){\makebox(0,0)[r]{\scriptsize Model}}
					\put(1,3){\makebox(0,0)[l]{\scriptsize $\Delta t\,(\mathrm{s})$}}
				\end{picture}
			} & PSTO & FNO & U-Net & U-FNO \\
			\midrule
			$10^{-6}$ & 6.94 & 1.74 & 1.84 & 2.08 \\
			$10^{-5}$ & 0.709 & 0.188 & 0.198 & 0.207 \\
			\bottomrule
		\end{tabular}
	\end{table}
	
	\section{Conclusion}
	\label{sec:conclusion}
	
	Coupled spatial transport and reactions make multiscale CDR simulations computationally expensive and their dynamics difficult to learn directly. PSTO addresses these challenges by learning transport from reaction-free CD trajectories and coupling the learned operator with independently pretrained local reaction models. Conditioned on velocity, density, and variable-specific transport coefficients, its shared transport block uses a trainable parameter count independent of the number of transported variables, enabling parameter-efficient modeling of high-dimensional CDR systems. Relaxed fixed-point iterations approximate a learned implicit transport update, and the accompanying theoretical analysis connects finite-iteration residuals and one-step approximation errors to error propagation over finite prediction horizons.
	
	Across the canonical CDR, Robertson counterflow, and hydrogen planar jet flame cases, PSTO reduces the mean variable-wise relative $L_2$ error after 100 coupled steps by at least $50\%$ compared with the best-performing neural baseline in each case. These improvements are achieved with fewer trainable transport parameters. For the planar jet flame, we compare PSTO trained and evaluated at $\Delta t=10^{-5}\,\mathrm{s}$ with all neural baselines at $\Delta t=10^{-6}\,\mathrm{s}$. Over the same physical horizon, PSTO achieves both a lower mean endpoint error and a shorter inference time. The corresponding wall-time speedup over the OpenFOAM workflow using 128 CPU MPI ranks is approximately 149-fold under the reported hardware configurations. These results highlight the potential of decoupled training and a shared implicit transport operator for efficient multiscale CDR modeling. Remaining errors in minor species and transverse velocity identify areas for improvement. Future work will extend the evaluation to longer prediction horizons and broader operating conditions, and investigate pressure coupling in three-dimensional CDR systems.
	
	\section*{Data availability}
    The source code and trained model weights associated with this study are publicly available at \url{https://github.com/kevin-sjtu/PSTO}. The repository also provides data-generation scripts, OpenFOAM case configurations, and instructions for data preparation.
	
	\section*{Declaration of competing interest}
	The authors declare no known competing financial or personal interests that could have appeared to influence the work reported in this paper.
	
	\section*{Declaration of generative AI and AI-assisted technologies in the manuscript preparation process}
	OpenAI Codex was used to assist with literature searches, mathematical exposition, coding and manuscript polishing. The authors take responsibility for the scientific claims, numerical results, citations and final manuscript content.
	
	\section*{Acknowledgments}
	This work is supported by the National Natural Science Foundation of China (92470104), AI for Science Program of the Shanghai Municipal Commission of Economy and Informatization (2025-GZL-RGZN-BTBX-02008), and Chinese Academy of Sciences Project for Young Scientists in Basic Research (YSBR-107).
	
	\appendix
	\numberwithin{equation}{section}
	\numberwithin{figure}{section}
	\numberwithin{table}{section}
	
	\section{Supplementary operator analysis}
	\label{app:proofs}
	
	\subsection{Proof of the fixed-point convergence and prediction-error bounds}
	\label{app:supplementary_residual_estimates}
	
	We first establish contraction of the relaxed latent update and then use this result to bound the transport prediction error. The input state, conditioning fields, and boundary data remain fixed throughout the iteration. Let $z$ and $z'$ denote latent fields with $K$ channels on the computational grid. All Euclidean inner products and norms are taken over the spatial grid and latent channels. The maps $P$ and $Q$ are applied pointwise and are interpreted as bounded linear operators on the corresponding discrete fields. Writing the transport block as $\mathcal T_\theta=A_\theta+D_\theta$, we define
	\[
	\Delta z=z-z',\qquad \Delta A=A_\theta(z)-A_\theta(z'),\qquad \Delta D=D_\theta(z)-D_\theta(z'),
	\]
	where the fixed conditioning arguments are omitted for brevity. Assume that, for some constants $m_F,\beta_F>0$ and all latent fields $z$ and $z'$,
	\begin{equation}
		\langle \Delta z,\Delta A+\Delta D\rangle \leq(1-m_F)\|\Delta z\|_2^2,\qquad \|\Delta z-\Delta A-\Delta D\|_2 \leq\beta_F\|\Delta z\|_2,\qquad 0<\omega <\min\left\{\tfrac12,\frac{2m_F}{\beta_F^2}\right\}.
		\label{eq:full_transport_conditions}
	\end{equation}
	Define the iteration map and update residual as
	\begin{equation}
		H_\theta(z;z^0) =z-\omega F_\theta(z;q^n,h),\qquad r_\ell =\|z^{\ell+1}-z^\ell\|_2 =\omega\|F_\theta(z^\ell;q^n,h)\|_2,
		\label{eq:iteration_map_residual}
	\end{equation}
	where $z^{\ell+1}=H_\theta(z^\ell;z^0)$ for $\ell\geq0$. The fixed arguments of $F_\theta$ and $H_\theta$ are suppressed below.
	
	\begin{proof}[Proof of Proposition~\ref{prop:iteration_fixed_point}]
		The difference between the residuals at $z$ and $z'$ is
		\[
		\Delta F =F_\theta(z)-F_\theta(z') =\Delta z-\Delta A-\Delta D.
		\]
		Expanding the squared residual difference retains the contributions of both branches and their interaction:
		\begin{equation}
			\|\Delta F\|_2^2 ={}\|\Delta z\|_2^2 -2\langle \Delta z,\Delta A\rangle -2\langle \Delta z,\Delta D\rangle +\|\Delta A\|_2^2 +\|\Delta D\|_2^2 +2\langle\Delta A,\Delta D\rangle.
			\label{eq:full_residual_expansion}
		\end{equation}
		The joint assumptions in Eq.~\eqref{eq:full_transport_conditions} imply
		\begin{equation}
			\langle \Delta z,\Delta F\rangle\geq m_F\|\Delta z\|_2^2, \qquad \|\Delta F\|_2\leq\beta_F\|\Delta z\|_2.
			\label{eq:complete_residual_monotonicity}
		\end{equation}
		Thus, $F_\theta$ is strongly monotone with constant $m_F$ and Lipschitz continuous with constant $\beta_F$. Consequently,
		\begin{equation}
			\|H_\theta(z)-H_\theta(z')\|_2^2 =\|\Delta z-\omega\Delta F\|_2^2 \leq (1-2\omega m_F+\omega^2\beta_F^2)\|\Delta z\|_2^2.
			\label{eq:full_transport_contraction}
		\end{equation}
		It follows that
		\begin{equation}
			\gamma =\sqrt{1-2\omega m_F+\omega^2\beta_F^2}<1,\qquad \|H_\theta(z)-H_\theta(z')\|_2 \leq\gamma\|z-z'\|_2.
			\label{eq:full_iteration_condition}
		\end{equation}
		The prescribed range of $\omega$ ensures $\gamma<1$. The Banach contraction theorem therefore yields a unique fixed point $z^*$ and convergence of the iterates to $z^*$. Since $\omega>0$, this fixed point is also the unique solution of $F_\theta(z)=0$.
		
		Applying contraction to successive iterates gives $r_\ell\leq\gamma r_{\ell-1}$ for $\ell\geq1$. Repeated application of this estimate and summation of the remaining increments yield
		\begin{equation}
			r_\ell \leq\gamma^\ell r_0,\qquad \|z^\ell-z^*\|_2 \leq\sum_{j=\ell}^{\infty}r_j \leq\frac{r_\ell}{1-\gamma}.
			\label{eq:fixed_point_bounds}
		\end{equation}
		
		At the lifted reference state $z^\dagger=Pq^{n+1}$, the definitions in Eq.~\eqref{eq:reference_residual_and_projection_error} give
		\[
		\|F_\theta(z^\dagger)\|_2=\epsilon_{\rm imp}, \qquad \|H_\theta(z^\dagger)-z^\dagger\|_2 =\omega\epsilon_{\rm imp}.
		\]
		Using contractivity and $H_\theta(z^*)=z^*$, we obtain
		\[
		\begin{aligned}
			\|z^*-z^\dagger\|_2 \leq \|H_\theta(z^*)-H_\theta(z^\dagger)\|_2 +\|H_\theta(z^\dagger)-z^\dagger\|_2 \leq \gamma\|z^*-z^\dagger\|_2 +\omega\epsilon_{\rm imp}.
		\end{aligned}
		\]
		Hence,
		\[
		\|z^*-z^\dagger\|_2 \leq\frac{\omega\epsilon_{\rm imp}}{1-\gamma}.
		\]
		Combining this estimate with Eq.~\eqref{eq:fixed_point_bounds} yields
		\[
		\|z^L-z^\dagger\|_2 \leq \|z^L-z^*\|_2+\|z^*-z^\dagger\|_2 \leq \frac{r_L+\omega\epsilon_{\rm imp}}{1-\gamma}.
		\]
		
		By the linearity of $P$ and $Q$, the physical prediction error can be decomposed as
		\[
		q_L-q^{n+1} = Q(z^L-z^\dagger) +(QP-I)(q^{n+1}-q^n).
		\]
		Taking norms and using the definition of $\epsilon_{PQ}$ proves Eq.~\eqref{eq:projected_implicit_accuracy}.
	\end{proof}
	
	For a fixed input, reference state, and trained model, Eq.~\eqref{eq:fixed_point_bounds} bounds the finite-iteration contribution to Eq.~\eqref{eq:projected_implicit_accuracy} by a geometrically decaying term.
	
	\subsection{Transport sensitivity and coupled error bounds}
	\label{app:proof_coupled_error}
	
	We first bound the transport and reaction approximation errors, propagate these bounds through the splitting compositions, and then estimate their accumulation during autoregressive prediction. We use Euclidean norms over all grid points and physical variables. For latent fields, the norms also include all feature channels. Operator norms are the corresponding induced Euclidean norms. All maps use the same state coordinates, and the exact flow maps refer to a fixed spatial discretization. Assume that all maps are well-defined and that all exact, approximate, and mixed intermediate states remain in an admissible set $\mathcal D$. The stated error and Lipschitz bounds are assumed finite and uniform on this set.
	
	The approximate counterparts of the splitting maps in Eq.~\eqref{eq:splitting} are
	\begin{equation}
		\widehat{\mathcal S}^{\rm LT}_{\Delta t}=\widehat{\mathcal R}_{\Delta t}\circ\widehat{\mathcal T}_{\Delta t},\qquad \widehat{\mathcal S}^{\rm St}_{\Delta t}=\widehat{\mathcal R}_{\Delta t/2}\circ\widehat{\mathcal T}_{\Delta t}\circ\widehat{\mathcal R}_{\Delta t/2},
		\label{eq:learned_and_exact_strang}
	\end{equation}
	where $\widehat{\mathcal T}_{\Delta t}$ denotes the PSTO transport update and $\widehat{\mathcal R}_\tau$ denotes the approximate reaction map over an interval $\tau$. For $\tau\in\{\Delta t/2,\Delta t\}$, define the uniform substep errors by
	\begin{equation}
		\varepsilon_{\mathcal T}(\Delta t)=\sup_{x\in\mathcal D}\|\widehat{\mathcal T}_{\Delta t}(x)-\mathcal T_{\Delta t}(x)\|_2,\qquad \varepsilon_{\mathcal R}(\tau)=\sup_{x\in\mathcal D}\|\widehat{\mathcal R}_\tau(x)-\mathcal R_\tau(x)\|_2.
		\label{eq:suboperator_errors}
	\end{equation}
	Let $\Lambda_{\mathcal T}(\Delta t)$, $\Lambda_{\mathcal R}(\tau)$, and $\Lambda_{\widehat{\mathcal R}}(\tau)$ denote Lipschitz constants of the exact transport, exact reaction, and approximate reaction maps on $\mathcal D$, respectively. Assume further that the exact splitting errors satisfy
	\begin{equation}
		\|\mathcal S^{\rm LT}_{\Delta t}(x)-\mathcal I_{\Delta t}(x)\|_2\leq C_{\mathrm L}\Delta t^2,\qquad \|\mathcal S^{\rm St}_{\Delta t}(x)-\mathcal I_{\Delta t}(x)\|_2\leq C_{\mathrm S}\Delta t^3,
		\label{eq:splitting_error}
	\end{equation}
	uniformly for $x\in\mathcal D$, with finite constants $C_{\mathrm L}$ and $C_{\mathrm S}$. 
	
	\begin{proof}[Proof of Proposition~\ref{prop:coupled_error_bound}]
		For the PSTO transport update, assume that Proposition~\ref{prop:iteration_fixed_point} applies uniformly to the full transport state for inputs in $\mathcal D$, with
		\[
		\gamma\leq\bar\gamma<1,\qquad r_L\leq\bar r_L,\qquad \epsilon_{\rm imp}\leq\bar\epsilon_{\rm imp},\qquad \epsilon_{PQ}\leq\bar\epsilon_{PQ}.
		\]
		Taking the supremum of Eq.~\eqref{eq:projected_implicit_accuracy} over $\mathcal D$ gives
		\begin{equation}
			\varepsilon_{\mathcal T}(\Delta t)\leq\bar\epsilon_{PQ} +\frac{\|Q\|_2}{1-\bar\gamma}\bigl(\bar r_L+\omega\bar\epsilon_{\rm imp}\bigr).
			\label{eq:transport_error_bridge}
		\end{equation}
		This estimate bounds the transport approximation error in terms of the reconstruction error, implicit residual, and remaining iteration residual. The implicit residual uses the conditioning fields supplied to PSTO, so constitutive-model error and coefficient freezing are included.
		
		For the reaction update, the MLP approximates the finite-time evolution generated by the coupled nonlinear kinetics $\dot{\xi}_R=g_R(\xi_R)$, where $\xi_R$ contains the local reacting variables. Let $s_R=E_R(\xi_R)$ denote the preprocessed input. The quantity $V_\tau^{\rm ex}(s_R)$ denotes the standardized increment associated with the exact reaction flow over $\tau$. The species networks can be stacked in the form
		\[
		a_R^0=s_R,\qquad a_R^\ell=\sigma_R(\mathbf W_\ell a_R^{\ell-1}+\mathbf b_\ell),\quad \ell=1,\ldots,d_R, \qquad N_{\theta_R,\tau}(s_R)=\mathbf W_{d_R+1}a_R^{d_R}+\mathbf b_{d_R+1}.
		\]
		Here $d_R$ is the number of hidden layers and $\sigma_R$ is the activation function. The affine transformations mix the input variables, while the nonlinear activations represent their coupled dependence. Thus, the approximation target is the nonlinear finite-time map $V_\tau^{\rm ex}$. For a continuous nonpolynomial activation, sufficiently expressive MLPs can approximate continuous maps uniformly on compact sets~\cite{leshno1993multilayer}.
		
		To quantify that error, let $\mathcal K_R$ be a compact set containing the preprocessed local inputs required during coupled prediction. Assume that $V_\tau^{\rm ex}$ is Lipschitz on $\mathcal K_R$, with constant $\Lambda_V(\tau)$. For an activation with Lipschitz constant $\Lambda_{\sigma_R}$, the network has a Lipschitz bound
		\[
		\Lambda_N(\tau)\leq \|\mathbf W_{d_R+1}\|_2\prod_{\ell=1}^{d_R} \bigl(\Lambda_{\sigma_R}\|\mathbf W_\ell\|_2\bigr).
		\]
		Given reference samples $\{(s_{R,i},V_i)\}_{i=1}^{M}$ with $s_{R,i}\in\mathcal K_R$, define
		\[
		\eta_{\rm fit}(\tau) =\max_i\|N_{\theta_R,\tau}(s_{R,i})-V_i\|_2, \qquad \eta_{\rm data}(\tau) =\max_i\|V_i-V_\tau^{\rm ex}(s_{R,i})\|_2, \qquad \varrho_M=\sup_{s_R\in\mathcal K_R}\min_i\|s_R-s_{R,i}\|_2.
		\]
		Here, $\eta_{\rm fit}$ is the maximum sample prediction error, $\eta_{\rm data}$ accounts for numerical error in the standardized targets, and $\varrho_M$ is the greatest distance from an admissible input to its nearest sample. The numerical reference integration is assumed sufficiently accurate that $\eta_{\rm data}$ is small relative to the surrogate approximation error. This term is nevertheless retained to account for differences between numerical and exact targets. For any $s_R\in\mathcal K_R$, choose a nearest sampled input $s_{R,i}$. The prediction error satisfies the exact identity
		\[
		\begin{aligned}
			N_{\theta_R,\tau}(s_R)-V_\tau^{\rm ex}(s_R) ={}&[N_{\theta_R,\tau}(s_R)-N_{\theta_R,\tau}(s_{R,i})] +[N_{\theta_R,\tau}(s_{R,i})-V_i]\\
			&+[V_i-V_\tau^{\rm ex}(s_{R,i})] +[V_\tau^{\rm ex}(s_{R,i})-V_\tau^{\rm ex}(s_R)].
		\end{aligned}
		\]
		The second and third terms are bounded by $\eta_{\rm fit}$ and $\eta_{\rm data}$. Applying the Lipschitz bounds to the first and fourth terms and using $\|s_R-s_{R,i}\|_2\leq\varrho_M$ gives
		\begin{equation}
			\eta_R(\tau) =\sup_{s_R\in\mathcal K_R} \|N_{\theta_R,\tau}(s_R)-V_\tau^{\rm ex}(s_R)\|_2 \leq \eta_{\rm fit}(\tau)+\eta_{\rm data}(\tau) +\bigl[\Lambda_N(\tau)+\Lambda_V(\tau)\bigr]\varrho_M.
			\label{eq:reaction_mlp_error}
		\end{equation}
		This estimate bounds the uniform prediction error using the maximum sample error, the numerical target error, and the distance to the nearest sampled input.
		
		The error in Eq.~\eqref{eq:reaction_mlp_error} is measured in standardized increment coordinates. To express it in physical variables, let $D_\tau(\xi_R,\zeta)$ apply the inverse scaling and transformations, update and normalize the species mass fractions, and recover temperature. Assume that reconstructing the physical state from the exact target reproduces the exact local reaction flow $r_\tau(\xi_R)$ and that this reconstruction map is uniformly Lipschitz continuous:
		\[
		D_\tau(\xi_R,V_\tau^{\rm ex}(E_R(\xi_R)))=r_\tau(\xi_R), \qquad \|D_\tau(\xi_R,\zeta)-D_\tau(\xi_R,\zeta')\|_2 \leq K_D(\tau)\|\zeta-\zeta'\|_2.
		\]
		These assumptions require consistent target encoding and well-conditioned
		physical reconstruction. In particular, species-normalization
		denominators and the mixture heat capacity used for temperature recovery
		must remain bounded away from zero. Since
		$\widehat r_\tau(\xi_R)=D_\tau(\xi_R,N_{\theta_R,\tau}(E_R(\xi_R)))$,
		the local physical error is at most $K_D(\tau)\eta_R(\tau)$.
		Summing its square over the $N_g$ spatial grid points therefore yields
		\begin{equation}
			\begin{aligned}
				\varepsilon_{\mathcal R}(\tau) &\leq \sqrt{N_g}\,K_D(\tau)\eta_R(\tau)\\
				&\leq \sqrt{N_g}\,K_D(\tau) \left[ \eta_{\rm fit}(\tau)+\eta_{\rm data}(\tau) +\bigl(\Lambda_N(\tau)+\Lambda_V(\tau)\bigr)\varrho_M \right].
			\end{aligned}
			\label{eq:reaction_error_bridge}
		\end{equation}
		The factor $\sqrt{N_g}$ arises from evaluating the Euclidean norm over all $N_g$ grid points. When a numerical reaction solver replaces the MLP, $\varepsilon_{\mathcal R}(\tau)$ denotes its uniform integration error over the reaction substep.
		
		For Lie--Trotter splitting, adding and subtracting $\widehat{\mathcal R}_{\Delta t}(\mathcal T_{\Delta t}(x))$ gives
		\[
		\|\widehat{\mathcal S}^{\rm LT}_{\Delta t}(x)-\mathcal S^{\rm LT}_{\Delta t}(x)\|_2 \leq \Lambda_{\widehat{\mathcal R}}(\Delta t)\varepsilon_{\mathcal T}(\Delta t)+\varepsilon_{\mathcal R}(\Delta t).
		\]
		Combining this estimate with Eq.~\eqref{eq:splitting_error} and taking the supremum over $\mathcal D$ yields
		\begin{equation}
			\delta_{\Delta t}^{\rm LT}\leq\varepsilon_{\mathcal R}(\Delta t) +\Lambda_{\widehat{\mathcal R}}(\Delta t)\varepsilon_{\mathcal T}(\Delta t)+C_{\mathrm L}\Delta t^2.
			\label{eq:complete_error_bound_lie}
		\end{equation}
		
		For Strang splitting, set $\tau=\Delta t/2$. After the first reaction and transport substeps, adding and subtracting $\mathcal T_{\Delta t}(\widehat{\mathcal R}_\tau(x))$ gives
		\[
		\|\widehat{\mathcal T}_{\Delta t}(\widehat{\mathcal R}_\tau(x))-\mathcal T_{\Delta t}(\mathcal R_\tau(x))\|_2 \leq\varepsilon_{\mathcal T}(\Delta t)+\Lambda_{\mathcal T}(\Delta t)\varepsilon_{\mathcal R}(\tau).
		\]
		At the final reaction substep, add and subtract $\mathcal R_\tau(\widehat{\mathcal T}_{\Delta t}(\widehat{\mathcal R}_\tau(x)))$. Applying the reaction approximation bound and the Lipschitz bound for $\mathcal R_\tau$, together with Eq.~\eqref{eq:splitting_error}, yields
		\begin{equation}
			\delta_{\Delta t}^{\rm St}\leq \bigl[1+\Lambda_{\mathcal R}(\tau)\Lambda_{\mathcal T}(\Delta t)\bigr]\varepsilon_{\mathcal R}(\tau) +\Lambda_{\mathcal R}(\tau)\varepsilon_{\mathcal T}(\Delta t)+C_{\mathrm S}\Delta t^3.
			\label{eq:complete_error_bound}
		\end{equation}
		Substituting Eqs.~\eqref{eq:transport_error_bridge} and~\eqref{eq:reaction_error_bridge} into these estimates bounds the transport and reaction contributions to the one-step error, in addition to the assumed splitting error.
		
		To estimate the sensitivity of PSTO to physical-input perturbations, consider the latent iterations $z^0(x)=Px$ and $z^{\ell+1}(x)=H_{\theta, x}(z^\ell(x))$, with
		\begin{equation}
			H_{\theta, x}(z)=(1-\omega)z+\omega\{Px+\mathcal T_\theta(z;h(x))\},\qquad \widehat{\mathcal T}_{\Delta t}(x)=(I-QP)x+Qz^L(x).
			\label{eq:physical_transport_map}
		\end{equation}
		Consider inputs $x,y\in\mathcal U\subseteq\mathcal D$ with the same prescribed boundary data. Assume that $H_{\theta, x}$ is uniformly contractive in its latent argument with factor $\bar\gamma<1$. Also assume that
		\begin{equation}
			\|H_{\theta, x}(z)-H_{\theta, y}(z)\|_2\leq\omega\kappa_{\mathcal U}\|x-y\|_2.
			\label{eq:conditioning_input_bound}
		\end{equation}
		Both bounds are assumed to hold on a common set containing all required latent iterates. Adding and subtracting $H_{\theta, x}(z^\ell(y))$ gives
		\[
		\|z^{\ell+1}(x)-z^{\ell+1}(y)\|_2 \leq\bar\gamma\|z^\ell(x)-z^\ell(y)\|_2+\omega\kappa_{\mathcal U}\|x-y\|_2.
		\]
		Since $z^0(x)-z^0(y)=P(x-y)$, repeated application of this inequality yields
		\[
		\|z^L(x)-z^L(y)\|_2 \leq\left[\bar\gamma^L\|P\|_2+\omega\kappa_{\mathcal U}\frac{1-\bar\gamma^L}{1-\bar\gamma}\right]\|x-y\|_2.
		\]
		Using the output map in Eq.~\eqref{eq:physical_transport_map}, we obtain
		$\|\widehat{\mathcal T}_{\Delta t}(x)-\widehat{\mathcal T}_{\Delta t}(y)\|_2\leq \Lambda_{\mathcal U,L}\|x-y\|_2$, where
		\begin{equation}
        \Lambda_{\mathcal U,L} = \|I-QP\|_2+\|Q\|_2 \left[ \bar\gamma^L\|P\|_2 +\omega\kappa_{\mathcal U} \frac{1-\bar\gamma^L} {1-\bar\gamma} \right].
        \label{eq:physical_transport_sensitivity}
    \end{equation}
		For PSTO, the local transport sensitivity $\Lambda_n^{\rm loc}$ in Fig.~\ref{fig:coupled_sensitivity_error}(b) satisfies $\Lambda_n^{\rm loc}\leq \Lambda_{\mathcal U,L}$ when both the reference and perturbed states lie in $\mathcal U$. This estimate accounts for changes in the input state and the corresponding conditioning fields.
		
		If $\Lambda_{\widehat{\mathcal T}}$ uniformly bounds transport sensitivity over all inputs required by the splitting maps, Lipschitz constants for the coupled maps can be chosen to satisfy
		\begin{equation}
			\Lambda_{\widehat{\mathcal S}^{\rm LT}}\leq \Lambda_{\widehat{\mathcal R}}(\Delta t)\Lambda_{\widehat{\mathcal T}},\qquad \Lambda_{\widehat{\mathcal S}^{\rm St}}\leq \Lambda_{\widehat{\mathcal R}}(\Delta t/2)^2\Lambda_{\widehat{\mathcal T}}.
			\label{eq:coupled_lipschitz_constant}
		\end{equation}
		Finally, adding and subtracting $\widehat{\mathcal S}^{\chi}_{\Delta t}(x_n)$ between the approximate and exact next states and dividing by $\|x_0\|_2$ gives
        \[
        e_{n+1} \leq \Lambda_{\widehat{\mathcal S}^{\chi}}e_n +\frac{\delta_{\Delta t}^{\chi}}{\|x_0\|_2}.
        \]
        Since the approximate and exact trajectories start from the same state, repeated application of this inequality gives
        \[
        e_n \leq \frac{\delta_{\Delta t}^{\chi}}{\|x_0\|_2} \sum_{j=0}^{n-1} \bigl(\Lambda_{\widehat{\mathcal S}^{\chi}}\bigr)^j,
        \]
        which proves Eq.~\eqref{eq:split_rollout_error}.
	\end{proof}

	\section{Initial and boundary conditions for the planar jet flame}
	\label{app:planarjet_setting}
	
\begin{figure}[pos=htbp]
		\centering
		\includegraphics[width=0.8\linewidth]{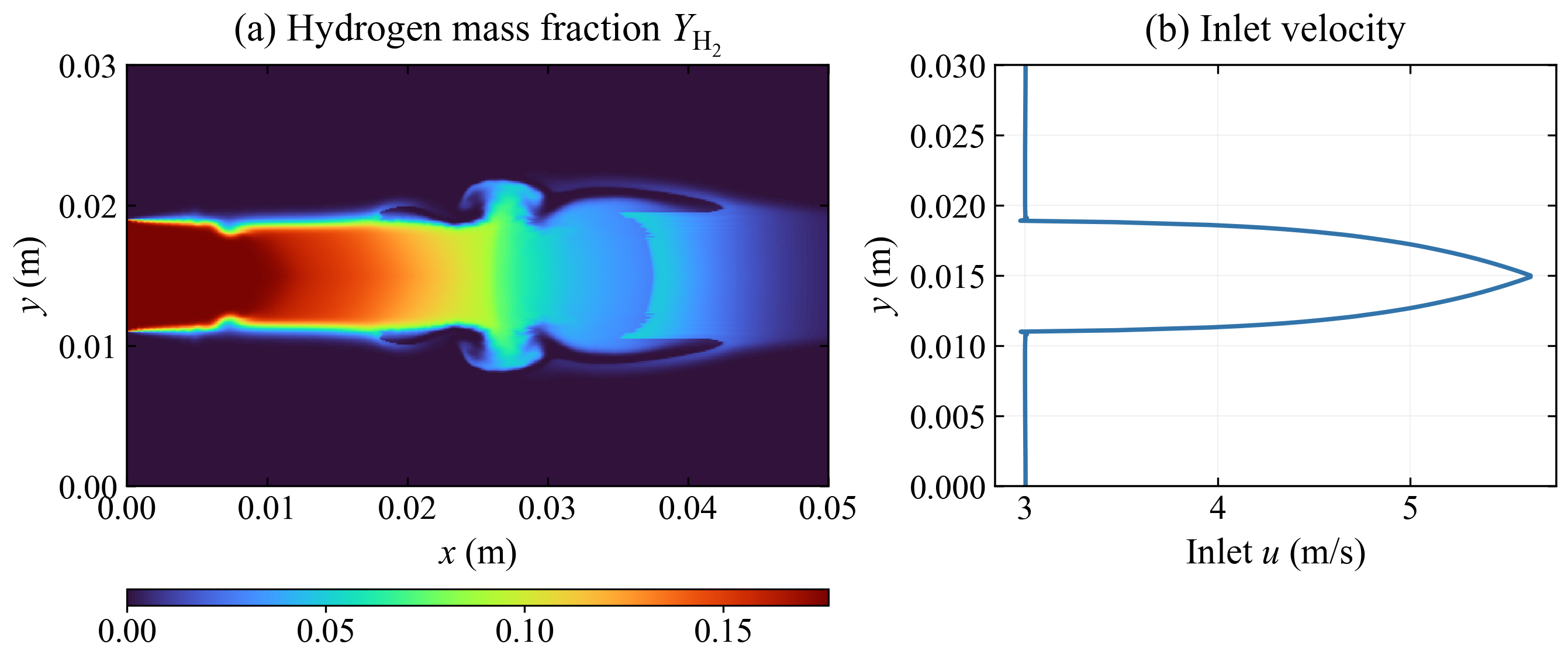}
		\caption{Initial and inlet conditions for the hydrogen planar jet flame. (a) Hydrogen mass fraction in the developed reacting state used to initialize coupled prediction. (b) Streamwise inlet velocity profile, with a faster central jet and slower air streams on both sides. The training and test compositions are specified in Section~\ref{sec:2d_planarjet}.}
		\label{fig:initial_h2_setting}
	\end{figure}

	This appendix specifies the reacting initial state and boundary treatment used for the predictions of planar jet flame in Section~\ref{sec:2d_planarjet}. Figure~\ref{fig:initial_h2_setting} illustrates the developed reacting state used to initialize coupled prediction and the prescribed inlet velocity profile. The streamwise and transverse coordinates are denoted by $x$ and $y$, respectively. At the inlet $x=0$, the central fuel stream occupies approximately $0.011\leq y\leq0.019\,\mathrm{m}$ and is centered at $y=0.015\,\mathrm{m}$, with air streams on either side. The prescribed central fuel-inlet profile is
	\begin{equation}
		u(y)=5\times1.128\left[1-\frac{|y-0.015\,\mathrm{m}|}{0.004\,\mathrm{m}}\right]^{1/7}\,\mathrm{m\,s^{-1}},\qquad 0.011\,\mathrm{m}\leq y\leq0.019\,\mathrm{m}.
		\label{eq:planarjet_inlet_profile}
	\end{equation}
	Its analytical maximum is $5.64\,\mathrm{m\,s^{-1}}$, and the sampled boundary maximum is $5.62987\,\mathrm{m\,s^{-1}}$. The adjacent air streams have a uniform inlet velocity of $3.0\,\mathrm{m\,s^{-1}}$. The profile shown in Fig.~\ref{fig:initial_h2_setting}(b) therefore represents the prescribed inlet configuration rather than downstream acceleration. The velocity difference produces two shear and mixing layers. Figure~\ref{fig:initial_h2_setting}(a) shows the hydrogen-rich central region and the downstream deformation of its interfaces with the surrounding air.
	
	Inlet values are enforced at each transport substep using the boundary values carried by its input state. Zero normal gradients are imposed at the downstream outlet, $x=0.05\,\mathrm{m}$, and at the transverse boundaries, $y=0$ and $y=0.03\,\mathrm{m}$. All surrogate models use identical initial and boundary data, including the treatment of corner values.
	
	The inlet configuration is based on the example from DeepFlame~\cite{mao2023deepflame}. The local case uses an initial temperature field with a maximum of $1200\,\mathrm{K}$, whereas the surrogate predictions begin from the subsequently developed reacting state shown in Fig.~\ref{fig:initial_h2_setting}.

	\section{Variable-wise prediction errors}
	\label{app:channel_errors}
	
	To clarify the contributions of individual variables to the mean errors reported in Section~\ref{sec:results}, Tables~\ref{tab:channels_CDR}--\ref{tab:channels_PlanarJet} list the variable-wise relative $L_2$ errors after 100 coupled steps, evaluated using Eq.~\eqref{eq:evaluation_metric}. PSTO achieves the lowest error for every reported variable across all three benchmarks.

    For the Robertson counterflow benchmark, PSTO reduces the errors in $Y_C$ and $Y_D$ by $94.6\%$ and $85.3\%$, respectively, relative to the best-performing baseline for each species. For the planar jet flame, the corresponding error reductions for the five minor species $\mathrm{H}$, $\mathrm{O}$, $\mathrm{OH}$, $\mathrm{HO_2}$, and $\mathrm{H_2O_2}$ range from $45.9\%$ to $61.0\%$.
    
    Table~\ref{tab:channels_PlanarJet_dt1e5} reports the variable-wise errors at $\Delta t=10^{-5}\,\mathrm{s}$ for the time-step sensitivity analysis in Section~\ref{sec:timestep_stability}. The transport surrogates and the shared reaction surrogate are trained for this larger time step. Ten coupled steps reach the same physical endpoint, $10^{-4}\,\mathrm{s}$, as the 100-step predictions in Table~\ref{tab:channels_PlanarJet}. At the larger time step, PSTO retains the lowest error for every variable among the neural surrogates, while its mean endpoint error remains nearly unchanged from the smaller-step result.
    
	\begin{table}[pos=htbp]
		\centering
		\caption{Variable-wise relative $L_2$ errors (\%) for the canonical CDR system after 100 coupled steps.}
		\label{tab:channels_CDR}
		\small
		\begin{tabular}{lrrrr}
			\toprule
			Variable & PSTO & FNO & U-Net & U-FNO \\
			\midrule
			$u$ & 1.44 & 5.64 & 3.26 & 4.25 \\
			$v$ & 1.10 & 2.28 & 2.12 & 2.53 \\
			$\phi$ & 0.136 & 0.553 & 0.723 & 0.692 \\
			\bottomrule
		\end{tabular}
	\end{table}
	
	\begin{table}[pos=htbp]
		\centering
		\caption{Relative $L_2$ errors (\%) in the four predicted species mass fractions for the Robertson counterflow benchmark after 100 coupled steps.}
		\label{tab:channels_Robertson}
		\small
		\begin{tabular}{lrrrr}
			\toprule
			Variable & PSTO & FNO & U-Net & U-FNO \\
			\midrule
			$Y_A$ & 0.371 & 6.90 & 3.85 & 3.57 \\
			$Y_B$ & 3.70 & 7.80 & 6.61 & 6.70 \\
			$Y_C$ & 1.11 & 20.5 & 26.2 & 29.8 \\
			$Y_D$ & 0.821 & 14.5 & 5.58 & 10.5 \\
			\bottomrule
		\end{tabular}
	\end{table}
	
	\begin{table}[pos=htbp]
		\centering
		\caption{Variable-wise relative $L_2$ errors (\%) in the twelve predicted fields for the planar jet flame after 100 coupled steps at $\Delta t=10^{-6}\,\mathrm{s}$, corresponding to a physical horizon of $10^{-4}\,\mathrm{s}$.}
		\label{tab:channels_PlanarJet}
		\small
		\begin{tabular}{lrrrr}
			\toprule
			Variable & PSTO & FNO & U-Net & U-FNO \\
			\midrule
			$T$ & 5.28 & 26.1 & 18.9 & 19.0 \\
			$Y_{\mathrm{H_2}}$ & 2.82 & 8.85 & 7.84 & 7.30 \\
			$Y_{\mathrm{H}}$ & 36.0 & 85.8 & 73.9 & 71.6 \\
			$Y_{\mathrm{O}}$ & 54.1 & 100 & 128 & 113 \\
			$Y_{\mathrm{O_2}}$ & 1.97 & 7.23 & 6.89 & 7.04 \\
			$Y_{\mathrm{OH}}$ & 38.3 & 88.7 & 88.4 & 83.8 \\
			$Y_{\mathrm{H_2O}}$ & 6.57 & 29.1 & 23.9 & 24.2 \\
			$Y_{\mathrm{HO_2}}$ & 48.4 & 108 & 96.9 & 101 \\
			$Y_{\mathrm{H_2O_2}}$ & 30.0 & 84.7 & 84.1 & 76.9 \\
			$Y_{\mathrm{N_2}}$ & 0.0805 & 0.712 & 0.453 & 0.289 \\
			$u$ & 7.68 & 9.51 & 9.50 & 9.53 \\
			$v$ & 48.4 & 67.5 & 67.3 & 67.4 \\
			\bottomrule
		\end{tabular}
	\end{table}
	
	\begin{table}[pos=htbp]
		\centering
		\caption{Variable-wise relative $L_2$ errors (\%) in the twelve predicted fields for the planar jet flame after 10 coupled steps at $\Delta t=10^{-5}\,\mathrm{s}$, corresponding to a physical horizon of $10^{-4}\,\mathrm{s}$. All four transport surrogates are coupled to the same reaction surrogate trained for this time interval.}
		\label{tab:channels_PlanarJet_dt1e5}
		\small
		\begin{tabular}{lrrrr}
			\toprule
			Variable & PSTO & FNO & U-Net & U-FNO \\
			\midrule
			$T$ & 6.37 & 29.7 & 22.1 & 32.1 \\
			$Y_{\mathrm{H_2}}$ & 6.63 & 11.4 & 7.34 & 35.8 \\
			$Y_{\mathrm{H}}$ & 41.9 & 314 & 131 & 130 \\
			$Y_{\mathrm{O}}$ & 50.5 & 106 & 93.1 & 100 \\
			$Y_{\mathrm{O_2}}$ & 2.02 & 12.5 & 10.3 & 9.18 \\
			$Y_{\mathrm{OH}}$ & 34.2 & 111 & 86.3 & 89.3 \\
			$Y_{\mathrm{H_2O}}$ & 8.49 & 42.2 & 31.7 & 46.5 \\
			$Y_{\mathrm{HO_2}}$ & 44.7 & 109 & 133 & 105 \\
			$Y_{\mathrm{H_2O_2}}$ & 30.0 & 83.4 & 94.4 & 82.0 \\
			$Y_{\mathrm{N_2}}$ & 0.0823 & 0.990 & 0.462 & 0.473 \\
			$u$ & 7.67 & 9.13 & 9.49 & 9.47 \\
			$v$ & 47.9 & 67.4 & 67.3 & 67.4 \\
			\bottomrule
		\end{tabular}
	\end{table}

	\section{Spatial error distributions}
	\label{app:spatial_errors}
	
	To locate the discrepancies underlying the variable-wise errors, we examine spatial error maps after 100 coupled steps for the same rollouts reported in Section~\ref{sec:results}. The displayed scalar is $\phi$ for canonical CDR, $Y_C$ for Robertson counterflow and $T$ for the planar jet flame. For each displayed variable $q_c$, the normalized pointwise absolute error is
	\begin{equation}
		e_c^{\rm spatial}(\mathbf x)=\frac{|\widehat q_c^{100}(\mathbf x)-q_c^{100}(\mathbf x)|}{\Delta q_{c,\max}^{\rm 100}},
		\label{eq:spatial_error}
	\end{equation}
	where $\Delta q_{c,\max}^{\rm 100}=q_{c,\max}^{100}-q_{c,\min}^{100}>0$ is the spatial range of the final reference field. All the maps within a case share the same error color scale, with scales set separately for each case. These maps complement the variable-wise relative $L_2$ errors by locating the spatial discrepancies.
	
	\begin{figure}[pos=htbp]
		\centering
		\includegraphics[width=0.5\linewidth]{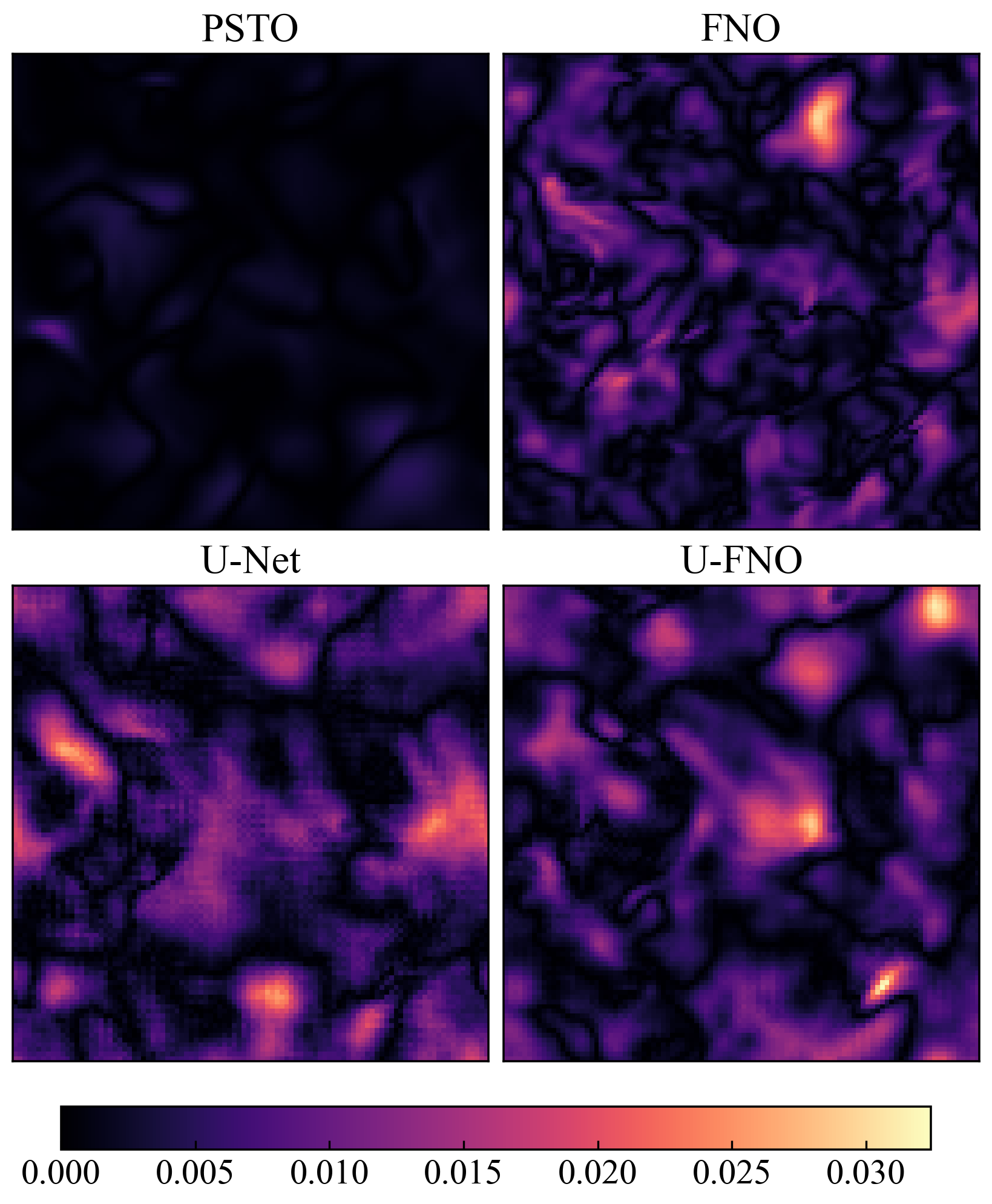}
		\caption{Normalized pointwise absolute error in $\phi$ for canonical CDR after 100 coupled steps [Eq.~\eqref{eq:spatial_error}]. All models use the same error color scale.}
		\label{fig:spatial_errors_cdr}
	\end{figure}
	
	\begin{figure}[pos=htbp]
		\centering
		\includegraphics[width=0.5\linewidth]{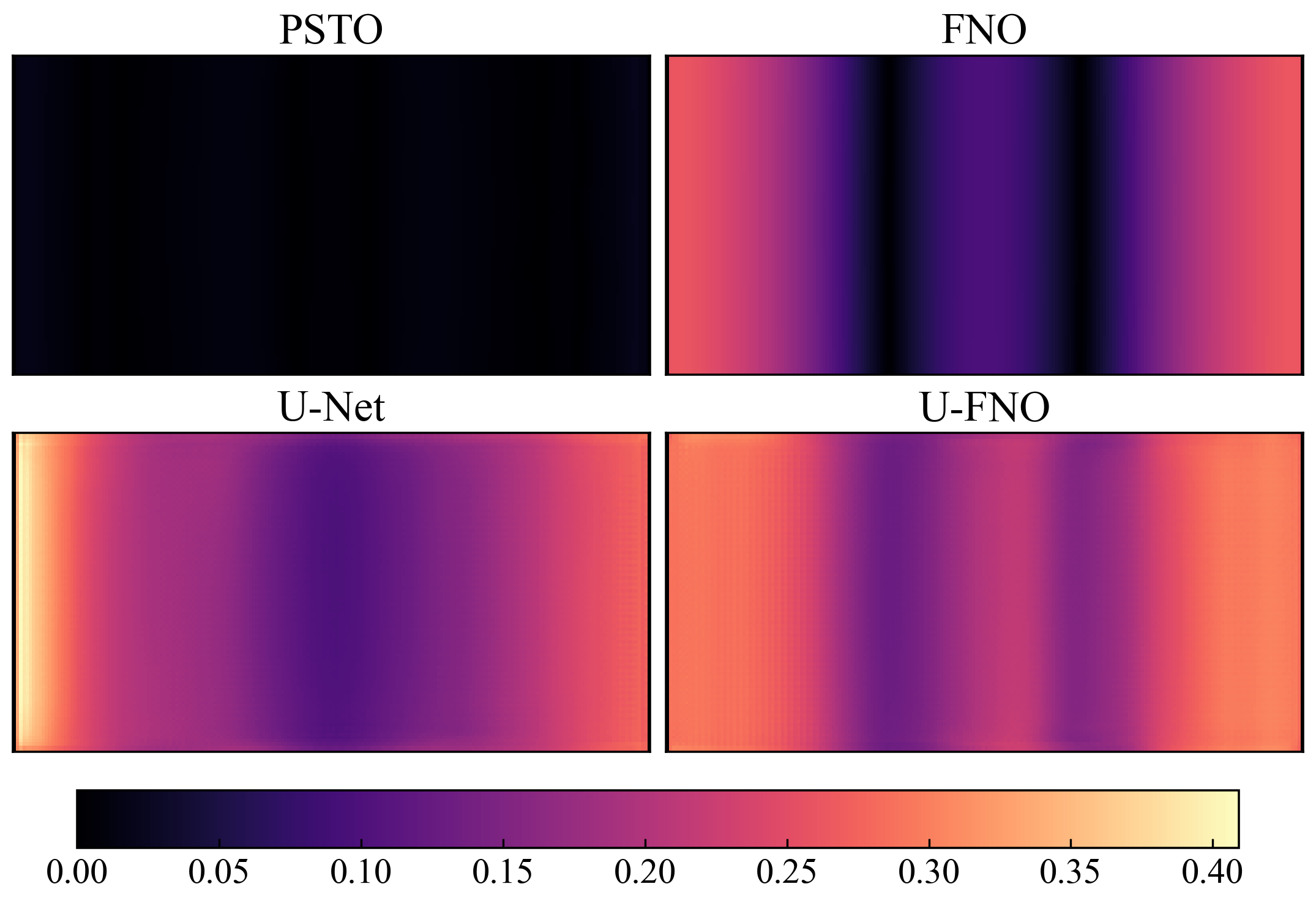}
		\caption{Normalized pointwise absolute error in $Y_C$ for Robertson counterflow after 100 coupled steps [Eq.~\eqref{eq:spatial_error}]. All models use the same error color scale.}
		\label{fig:spatial_errors_robertson}
	\end{figure}
	
	\begin{figure}[pos=htbp]
		\centering
		\includegraphics[width=0.5\linewidth]{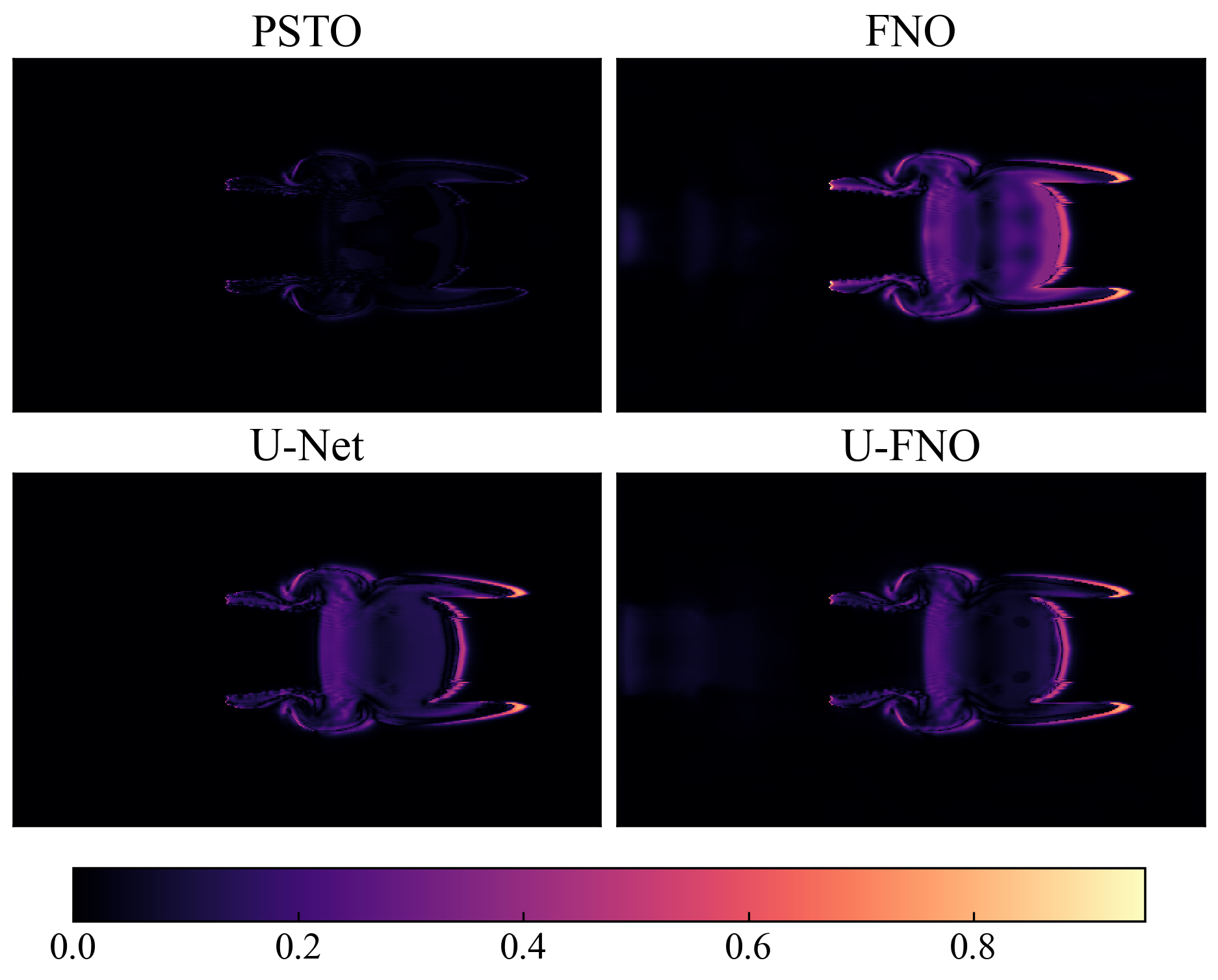}
		\caption{Normalized pointwise absolute error in $T$ for the planar jet flame after 100 coupled steps [Eq.~\eqref{eq:spatial_error}]. All models use the same error color scale.}
		\label{fig:spatial_errors_jet}
	\end{figure}
	
	Figure~\ref{fig:spatial_errors_cdr} shows smaller and more localized scalar errors for PSTO. In Robertson counterflow, PSTO reduces the widespread discrepancies in the product mass fraction $Y_C$ observed for the baselines [Fig.~\ref{fig:spatial_errors_robertson}]. In the planar jet flame, it reduces temperature errors in the downstream hot region and along the curved flame layers and their tips [Fig.~\ref{fig:spatial_errors_jet}].

	\section{Model configurations, training settings and validation histories}
	\label{app:training_settings}
	
	This appendix documents the model configurations, reaction-solver settings, training protocols, and validation histories used in the numerical comparisons. Table~\ref{tab:model_sizes} lists the transport architecture settings and parameter counts. PSTO contains 197,841 trainable parameters, compared with approximately 1.64--3.04 million for the baselines, corresponding to a factor of 8.3--15.4.
	
	\begin{table}[pos=htbp]
		\centering
		\caption{Transport architecture settings and trainable parameter counts. The real and imaginary components of complex Fourier weights are counted separately. Frozen constitutive and reaction networks are excluded.}
		\label{tab:model_sizes}
		\small
		\setlength{\tabcolsep}{4pt}
		\begin{tabular}{llrrrr}
			\toprule
			Case & Model & Parameters & $m$ & $w_{\rm net}$ & $L_{\rm arch}$ \\
			\midrule
			Canonical CDR & PSTO & 197,841 & 16 & 8 & 4 \\
			& FNO & 1,640,223 & 16 & 20 & 4 \\
			& U-Net & 3,037,743 & --- & 20 & 4 \\
			& U-FNO & 2,013,183 & 16 & 20 & 4 \\
			\midrule
			Robertson counterflow & PSTO & 197,841 & 16 & 8 & 4 \\
			& FNO & 1,640,304 & 16 & 20 & 4 \\
			& U-Net & 3,037,824 & --- & 20 & 4 \\
			& U-FNO & 2,013,264 & 16 & 20 & 4 \\
			\midrule
			Planar jet flame & PSTO & 197,841 & 16 & 8 & 4 \\
			& FNO & 1,640,592 & 16 & 20 & 4 \\
			& U-Net & 3,038,112 & --- & 20 & 4 \\
			& U-FNO & 2,013,552 & 16 & 20 & 4 \\
			\bottomrule
		\end{tabular}
		\par\medskip
		\begin{minipage}{\linewidth}
			\footnotesize
			The parameter $m$ denotes the number of retained Fourier modes per spatial direction. For PSTO, $w_{\rm net}=K$ denotes the number of latent channels per physical variable, and $L_{\rm arch}=L$ denotes the shared iteration count. For the baselines, $w_{\rm net}$ denotes the hidden width. The parameter $L_{\rm arch}$ counts Fourier blocks in FNO and U-FNO and encoder levels in U-Net. Dashes indicate inapplicable settings.
		\end{minipage}
	\end{table}
	
	For Robertson counterflow surrogate prediction, the common reaction solver uses adaptive SDIRK2 integration in \texttt{float64} arithmetic, with a relative tolerance of $10^{-8}$ and absolute tolerances of $10^{-13}$ for $Y_A$ and $10^{-17}$ for $Y_B$. $Y_C$ is reconstructed from conservation of $Y_A+Y_B+Y_C$, and inert $Y_D$ remains unchanged.
	
	Table~\ref{tab:training_settings} gives the chronologically ordered training and validation partitions and the number of training epochs. Tables~\ref{tab:optimizer_settings} and~\ref{tab:loss_settings} list the common optimizer settings and case-specific loss weights, respectively. Within each case, all four transport surrogate models use the same data partitions, optimization schedule and loss weights. One epoch corresponds to a complete pass through the training pairs.
	
	\begin{table}[pos=htbp]
		\centering
		\caption{Reaction-free CD datasets and training schedules for the four transport surrogate models.}
		\label{tab:training_settings}
		\small
		\begin{tabular}{lrrrl}
			\toprule
			Case & Training pairs & Validation pairs & Epochs & Learning-rate schedule \\
			\midrule
			Canonical CDR & 1600 & 400 & 100 & Linear warmup + cosine decay \\
			Robertson counterflow & 4000 & 1000 & 40 & Linear warmup + cosine decay \\
			Planar jet flame & 4000 & 1000 & 40 & Linear warmup + cosine decay \\
			\bottomrule
		\end{tabular}
	\end{table}
	
	\begin{table}[pos=htbp]
		\centering
		\caption{Optimizer settings shared across the three cases.}
		\label{tab:optimizer_settings}
		\small
		\begin{tabular}{ll}
			\toprule
			Setting & Value \\
			\midrule
			Optimizer & AdamW \\
			Initial learning rate & $3\times10^{-4}$ \\
			Final learning rate & $10^{-6}$ \\
			Warmup & 100 optimizer updates, linear \\
			Schedule after warmup & Cosine decay \\
			Weight decay & $10^{-5}$ on nonspectral parameters \\
			Gradient clipping & Unit norm \\
			\bottomrule
		\end{tabular}
	\end{table}
	
	\begin{table}[pos=htbp]
		\centering
		\caption{Case-specific loss weights shared by all four transport surrogate models.}
		\label{tab:loss_settings}
		\small
		\setlength{\tabcolsep}{4pt}
		\begin{tabular}{lrrrrr}
			\toprule
			Case & $\lambda_{\rm ROI}$ & $\lambda_{\rm wave}$ & $\lambda_{\rm PDE}$ & $\lambda_{\rm mass}$ & $\lambda_{\rm neg}$ \\
			\midrule
			Canonical CDR & 0.1 & 0.02 & 0.1 & --- & --- \\
			Robertson counterflow & 0.1 & 0.02 & 0.1 & 1 & 1 \\
			Planar jet flame & 0.1 & 0.02 & 0.1 & 1 & 1 \\
			\bottomrule
		\end{tabular}
		\par\medskip
		\begin{minipage}{\linewidth}
			\footnotesize
			All models use ROI exponent $o=1$, with weights computed from $\phi$ for canonical CDR, $Y_B$ for Robertson counterflow, and $T$ for the planar jet flame. The wavelet settings and adaptive PDE factor follow Eqs.~\eqref{eq:wavelet_loss} and~\eqref{eq:psto_total_loss}. Dashes indicate inapplicable penalties.
		\end{minipage}
	\end{table}
	
	\begin{figure}[pos=htbp]
		\centering
		\includegraphics[width=0.98\linewidth]{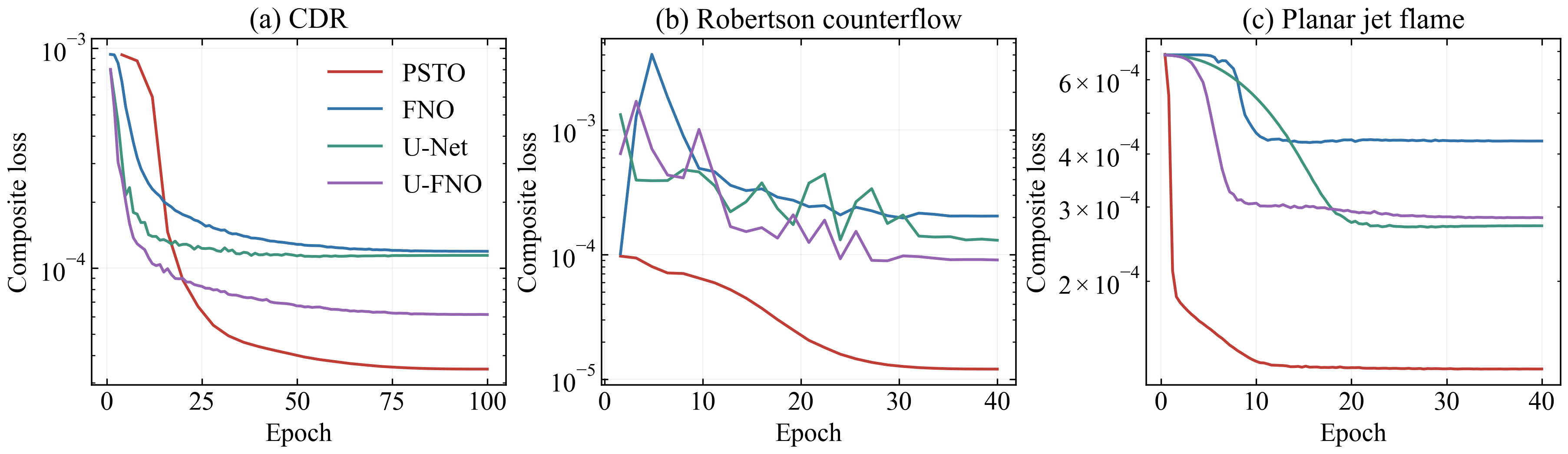}
		\caption{Reaction-free CD validation loss versus training epoch for (a) canonical CDR, (b) Robertson counterflow and (c) the planar jet flame. The checkpoint with the lowest validation loss is selected for coupled evaluation. Loss values depend on the fields and applicable loss terms in each case.}
		\label{fig:training_curves}
	\end{figure}
	
	Figure~\ref{fig:training_curves} shows the validation loss histories used for checkpoint selection. For canonical CDR, PSTO reaches the lowest validation loss by the end of training. In Robertson counterflow, its validation loss decreases smoothly and remains below those of the baselines. For the planar jet flame, PSTO rapidly reaches and maintains a low validation loss. Checkpoints are selected using the reaction-free CD validation histories. The selected transport surrogates are then evaluated on CDR trajectories without further training.
	
	\FloatBarrier
	\bibliographystyle{elsarticle-num}
	\bibliography{cas-refs}
\end{document}